\documentclass[letter,10pt,conference]{ieeeconf}

\IEEEoverridecommandlockouts
\usepackage[utf8]{inputenc}
\usepackage{times}
\usepackage{multicol}
\usepackage{graphicx}
\usepackage{amsmath}
\usepackage{amssymb}
\usepackage{bm}
\usepackage{mathtools}
\usepackage[font=footnotesize]{caption,subfig}
\usepackage{booktabs}
\usepackage{multirow}
\usepackage{tikz}
\usetikzlibrary{positioning, calc, decorations.pathreplacing}
\usepackage{pgfplots}
\usepackage{algorithm}
\usepackage[noend]{algpseudocode}
\usepackage{float}
\floatstyle{plaintop}
\restylefloat{table}

\makeatletter
\let\NAT@parse\undefined
\makeatother

\usepackage[bookmarks=true]{hyperref}
\definecolor{blueLink}{rgb}{0.05,0.05,0.6}
\hypersetup{  
    colorlinks=true,
    linkcolor=blueLink,
    filecolor=blueLink,
    urlcolor=blueLink,
    citecolor=blueLink
}

\newcommand{\AgilusBig}{KUKA Agilus KR~10}
\newcommand{\AgilusSmall}{KUKA Agilus KR~6}

\newcommand{\DatasetLin}{LINEAR}

\newcommand{\DatasetSpline}{SPLINE}

\newcommand{\AlgoNN}{(GCC-NN)}

\newcommand{\AlgoGCC}{(GCC$_0$)}

\newcommand{\grasp}{$\textbf{q}_\text{grasp}$}
\newcommand{\preStart}{$\textbf{q}_\text{start}$}
\newcommand{\preVia}{$\textbf{q}_\text{inVia}$}
\newcommand{\postVia}{$\textbf{q}_\text{outVia}$}
\newcommand{\postEnd}{$\textbf{q}_\text{end}$}

\newcommand{\motionPre}{$\textbf{m}_\text{in}$}
\newcommand{\motionPost}{$\textbf{m}_\text{out}$}

\begin{document}

\title{\LARGE \bf Neural Implicit Swept Volume Models for Fast Collision
  Detection}

\author{\authorblockN{Dominik Joho$^{*}$ \qquad Jonas Schwinn$^{*}$ \qquad Kirill Safronov%
    \thanks{$^{*}$ Equal contribution. \newline
    The authors are with KUKA Deutschland GmbH, Germany.
    \{Dominik.Joho, Jonas.Schwinn, Kirill.Safronov\}@kuka.com}}%
\authorblockA{}
}

\maketitle
\thispagestyle{empty}
\pagestyle{empty}

\begin{abstract}
  Collision detection is one of the most time-consuming operations
  during motion planning. Thus, there is an increasing interest in
  exploring machine learning techniques to speed up collision
  detection and sampling-based motion planning. A recent line of
  research focuses on utilizing neural signed distance functions of
  either the robot geometry or the swept volume of the robot motion.
  Building on this, we present a novel neural implicit swept volume
  model to continuously represent arbitrary motions
  parameterized by their start and goal configurations. This allows to
  quickly compute signed distances for any point in the task space to
  the robot motion. Further, we present an algorithm combining the
  speed of the deep learning-based signed distance computations with
  the strong accuracy guarantees of geometric collision checkers. We
  validate our approach in simulated and real-world robotic
  experiments, and demonstrate that it is able to speed up a
  commercial bin picking application.
\end{abstract}

\section{Introduction}

Bin picking is a crucial task in logistics and industrial automation processes. 
Solving this task typically involves several steps, such as object localization, grasp pose selection, 
and collision-free motion planning~\cite{kleeberger2020crr},~\cite{tenpas2017ijrr}. 
There exist multiple paradigms for finding a collision-free path: 
for example, one can use a path planer (sampling-based or optimization-based) to construct a single collision-free path. 
In practice, bin picking applications often follow a different paradigm where a set of paths is first heuristically generated
and subsequently checked for collisions until a first collision-free path has been found.

Besides object detection, these collision checks are one of the most time-consuming steps in bin picking applications. 
Geometric collision detection methods typically discretize the motion along a path into small-enough steps and check the resulting 
robot configurations against the environment~\cite{pan15engineering,coleman14joser}. In this paper, we follow a different approach utilizing the 
so-called swept volume of a robot motion, which is the subset of the workspace that the robot moves through during this motion. 
An explicit representation of this volume would allow to replace the individual collision 
checks along a path segment with a single collision check of this swept volume against the environment.

\begin{figure}[t]
  \centering
  \includegraphics[width=0.75\columnwidth]{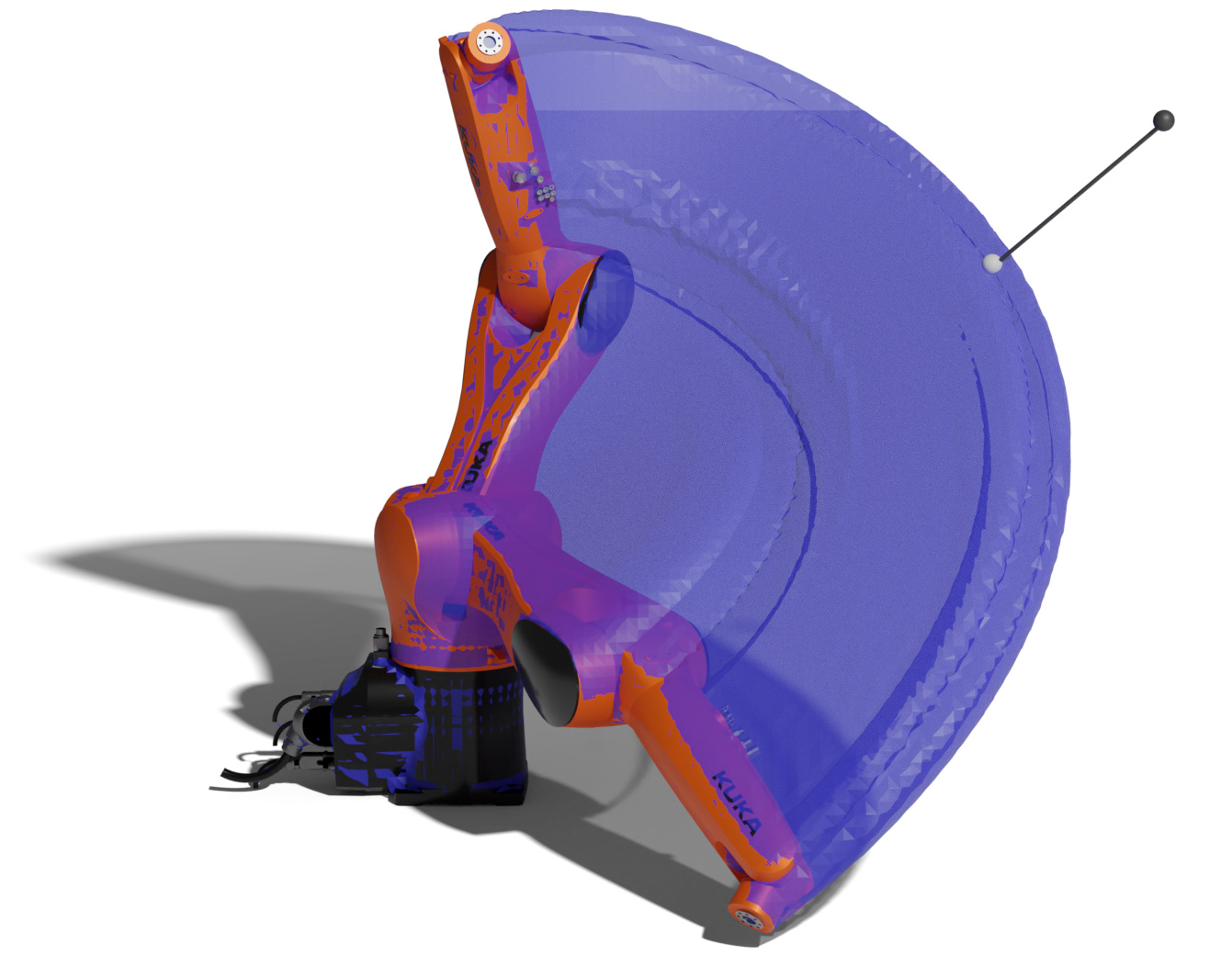}
  \caption{The basic idea of our approach is to predict the signed
    distance (black line) of a point in the task space (black dot) to
    its nearest point (white dot) on the surface of the swept volume
    (blue mesh) of a robot motion. The swept volume is implicitly
    represented by a neural network, which takes the start and goal
    configuration of the motion as well as the query point as inputs
    and outputs the signed distance of the point to the implied swept
    volume. The robot motion can be collision checked against a
    complete environment, by representing the environment as a point
    cloud. Each of these points becomes one input point for the
    network.}
  \label{fig:intro}
\end{figure}

However, computing an \emph{explicit} representation of the swept
volume comes with its own drawbacks. It is usually done via voxel grid
approximations, or by geometrically approximating the boundary of this
volume, e.g., as a triangle mesh~\cite{sellan21tog}. By design the
accuracy of these approaches is strongly linked to their computation
time, which hinders their application for collision checking in live
systems.

We thus propose an \emph{implicit} representation based on a deep neural network that learns 
the function $(\mathbf{x}, \mathbf{q}_0, \mathbf{q}_1) \mapsto \delta$, where $\mathbf{q}_0$ is the start configuration of the robot, 
$\mathbf{q}_1$ is the goal configuration, $\mathbf{x} \in \mathbb{R}^3$ is the query point in the task space for which we want 
to perform a collision check, and $\delta$ is the signed distance of the query point to the resulting swept volume when performing the motion $\mathbf{q}_0 \rightarrow \mathbf{q}_1$, see also Fig.~\ref{fig:intro}.
This approach learns a continuous implicit swept volume model, i.e., it can predict a signed distance for any point in the task space, 
and is not limited to a fixed resolution of a voxel grid. Further, for a given network the inference time is independent of the
length of the motion.

This approach can be used to check a robot motion against a complete
environment represented as a point cloud. Each point $\mathbf{x}$ of
this point cloud becomes one input for the network. Multiple points
can be batch processed leveraging parallel computation on GPUs, which
allows for favorable computation times compared to geometric collision
checkers (GCC). However, a downside of this approach is that the
distance computed by a neural network is less reliable when compared
to a GCC. Neural networks cannot be guaranteed to extrapolate robustly
to data not seen during training~\cite{kaidi2019neurips},
\cite{tan2020cvpr}, \cite{wang2020iclr}. This hinders applications in
industrial settings, as trading off accuracy against speed is not
acceptable when it comes to safety and potential collisions.

We therefore propose an algorithm that interleaves our implicit swept
volume model with a GCC. As we will show in
our experiments, this effectively reduces the number of paths that
have to be checked by the GCC and thus also
reduces the total time spent to find a collision-free path, even when
considering the additional time spent on computing the swept volume
distances.

While our primary example revolves around a bin picking application,
it is worth noting that the same model is also suitable for collision
checks in general sampling-based motion planners. Moreover, while our
evaluation focuses on a robotic arm, our approach can be broadly
applied to any articulated object that allows for swept volume
computation.

The main contributions of this paper are: (a) a novel approach to
modeling robot motions as implicit swept volumes parameterized by the
start and goal configuration with a neural network, and (b) an
algorithm that utilizes these neural swept volumes to speed up
collision checking while preserving accuracy guarantees of GCCs.

\section{Related Work}

Chiang et~al.~\cite{chiang20wafr} proposed a neural network that
learned to map the start and goal configuration of a robot motion to
the swept volume (as a scalar value) of this motion. The motion is
assumed to be linear in joint space. This is not usable for collision
detection, but has been used as a heuristic to speed up a
sampling-based motion planner. Building on this, Baxter
et~al.~\cite{baxter20iros} proposed a network that uses the same
input, but outputs a voxel grid representation of the swept volume
(voxels can either be ``swept'' or ``not swept''). This method can be
used for collision checks, but on the downside it has an inherent
discretization error determined by the size of the voxel grid cells,
and a low spatial resolution due to the limited number of voxels. In
contrast, in a feasibility study Lee et~al.~\cite{lee22ifac} used the
deep signed distance model of Gropp et~al.~\cite{gropp20icml} to model
swept volumes implicitly as a signed distance field (SDF), thereby
avoiding discretizations. However, their model is not conditioned on
motion parameters, but directly maps a point in the task space to the
signed distance of an implicit swept volume. Thus, each trained
network represents the swept volume of only one specific motion.
Michaux et~al.~\cite{michaux23rss} learn a parameterized signed
distance function of zonotope approximations of a box obstacle and a
swept volume of a robot motion. The inputs of their network are the
box obstacle center, as well as the initial configuration, speed, and
acceleration of the robot, which uniquely determines the motion of a
robot. They use this SDF as a collision avoidance constraint in receding
time horizon planning.

Besides modeling swept volumes, which represent complete motions, deep
neural networks have also been used in motion planning for modeling
the SDF of the robot geometry at a given joint configuration. Liu
et~al.~\cite{liu22iros} learned a SDF for articulated objects. Their
model takes as input a 3D query point and the configuration of the
articulated object (robot or human), and outputs the signed distance.
They used the model in motion planning for computing repulsive
velocities. Koptev et~al.~\cite{koptev23ral} presented a model that,
given a joint configuration, computes the minimal distance for each
link of a robot to a given 3D query point. They use the gradients of
the distances for collision avoidance in their control loop. Li
et~al.~\cite{li23signedDist} proposed to learn a SDF for each link
based on basis functions and ridge regression and compare their
approach to deep learning-based methods.

In contrast to Baxter et~al.~\cite{baxter20iros}, our model is a SDF
which can continuously represent the distance function and thereby
avoids discretization errors and the computational burden of the
network having to output a dense 3D voxel grid. In contrast to Lee
et~al.~\cite{lee22ifac}, our model is conditioned on the motion
parameters, and thus the trained network is generally applicable to
any motion of the robot. Closest to our implicit swept volume model is
the work of Michaux et~al.~\cite{michaux23rss}. They use a specific
motion type in which the motion is partitioned in one phase with
constant acceleration, followed by a second phase with constant
deceleration. Their swept volumes are parameterized by the initial
configuration, speed, and acceleration of the motion. We allow for
arbitrary motions parameterized by their start and goal configuration.
Further, we represent the obstacles nonparametrically as point clouds,
while they use zonotope approximations for the swept volume and the
box obstacles.

\section{Implicit Swept Volume Model}

The main idea of our approach is to compute the signed distance of a
point in the task space to the swept volume of a robot motion. If the
signed distance is positive (or greater than a given safety margin),
we would consider this motion as collision-free w.r.t. the given point.
To check the robot motion against the complete environment,
we represent the environment as a point cloud and check the
individual points against the swept volume.

We are thus interested in a function $f_d$ that outputs the signed
distance of a given point $\mathbf{x} \in \mathbb{R}^3$ in the task
space to the swept volume $V_S(\mathbf{m}) \subset \mathbb{R}^3$ of
a given motion $\mathbf{m}$:
\begin{equation}
  f_d\colon (\mathbf{x}, \mathbf{m}) \mapsto d(\mathbf{x}, V_S(\mathbf{m})),
  \label{eq:fd}
\end{equation}
where $d(\mathbf{x}, V)$ is the signed distance of a point
$\mathbf{x}$ to the boundary of some volume $V$. The motion is
considered collision-free if $f_d(\mathbf{x}, \mathbf{m}) > 0$. Here,
$\mathbf{m}$ are motion parameters that suffice to uniquely define a
concrete motion of a certain motion type, e.g., a linear motion in
joint space parameterized by the start and goal configurations of the
robot: $\mathbf{m} = (\mathbf{q}_0, \mathbf{q}_1)$, where
$\mathbf{q}_0$ and $\mathbf{q}_1 \in \mathbb{R}^{\text{dof}}$ are
joint configurations and $\text{dof}$ is the degrees of freedom of the
robot. We can denote the motion of the robot in the joint space as
\begin{equation}
  c \colon (\mathbf{m}, s) \mapsto \mathbf{q}_s \qquad s\in [0,1]
\end{equation}
where the motion is supposed to start at $s=0$ and end at $s=1$. If we
denote by $V_R(\mathbf{q}_s) \subset \mathbb{R}^3$ the volume in the
task space occupied by the robot at the joint configuration
$\mathbf{q}_s$, then the swept volume can be defined as
\begin{align}
  V_S(\mathbf{m}) &=  \bigcup_{s \in [0,1]} V_R(c(\mathbf{m}, s)) \\
                    &= \bigcup_{s \in [0,1]} V_R(\mathbf{q}_s).
\end{align} 

Computing an explicit representation that approximates the swept
volume $V_S$, e.g., as a triangle mesh, and using this representation
to compute the signed distance to the point $\mathbf{x}$ would be too
slow to be useful in motion planning. We therefore learn an implicit
representation of the swept volume as a signed distance field, by
training a deep neural network that directly computes the function
$f_d$.

\section{Method}

In the following, we describe the network architecture, the
integration with geometric collision checking, and the generation of
training data.

\subsection{Network Architecture}

\begin{figure}[t]
  \centering
  \vspace{2ex}
  \includegraphics[width=0.95\columnwidth]{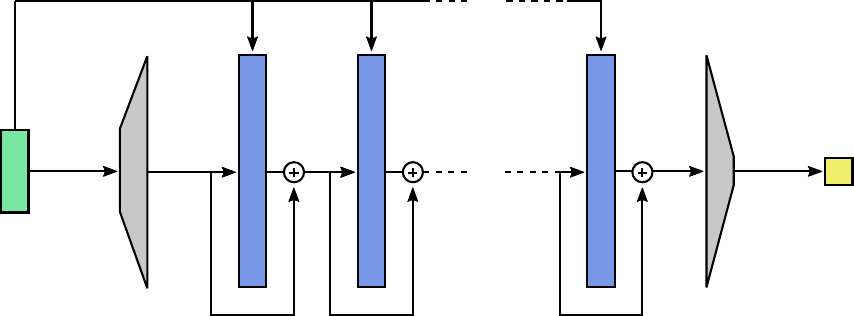}
  \caption{Network architecture: each blue rectangle represents one of
    the $n_{\text{b}}$ blocks of the network. The input (green) of the
    network is fed into each block individually. The gray layers are
    linearly mapping the input and output dimension of the network to
    the constant block dimension $n_{\text{dim}}$.}
  \label{fig:net}
\end{figure}

We use a simple, fully connected neural network $f_{\bm{\theta}}$ with
parameters $\bm{\theta}$ to approximate the function $f_d$ of
(\ref{eq:fd}). The main component of our network is a block consisting
of a sequence of layers. These are a linear layer with constant input
and output dimension, a ReLU activation function, and a BatchNorm
layer. We introduce skip connections~\cite{resnet} for each block
adding the block input to the block output. The block dimension
$n_\text{dim}$ denotes the input (and output) dimension of its linear
layer. Our networks are composed of $n_\text{b}$ blocks of equal
dimension $n_\text{dim}$. These are the two most important
hyperparameters of the architecture. Inspired by~\cite{park19cvpr}, we
further introduce jump connections, concatenating the input of the
network to each output of a block. Finally, the blocks are enclosed in
an initial linear layer mapping from the input dimension to the block
dimension, and a final layer mapping the block dimension to a single
scalar value, the predicted signed distance. The architecture of the
network is depicted in Fig.~\ref{fig:net}.

\subsection{Integration with Geometric Collision Checking}

We address the problem of finding a collision-free motion in a
predefined set of motions $\mathbf{M} = \{\mathbf{m}_j\}_{j=1}^{m}$.
Given a point cloud
$\mathbf{X} = \{\mathbf{x}_i \in \mathbb{R}^3\}_{i=1}^{n}$ of the
environment and some motion $\mathbf{m} \in \mathbf{M}$, we are
interested in the minimum distance of the point cloud to the swept
volume
\begin{equation}
  d(\mathbf{X}, \mathbf{m}) \coloneqq \min\limits_{\mathbf{x} \in \mathbf{X}} d(\mathbf{x}, V_S(\mathbf{m})).
\end{equation}
To find a collision-free motion in $\mathbf{M}$, we like to find some
$\mathbf{m} \in \mathbf{M}$ satisfying
$d(\mathbf{X}, \mathbf{m}) > \epsilon_{S}$, where $\epsilon_{S}$ is a
safety margin set by the user. Computation of
$d(\mathbf{X}, \mathbf{m})$ for each $\mathbf{m} \in \mathbf{M}$ by a
GCC is time-consuming. Therefore, we
make use of our neural implicit swept volume model
$f_{\bm{\theta}}(\cdot, \cdot)$ and compute
\begin{equation}
  \hat{d}(\mathbf{X}, \mathbf{m}) \coloneqq \min\limits_{\mathbf{x} \in \mathbf{X}} f_{\bm{\theta}}(\mathbf{x}, \mathbf{m})
\end{equation}
as a proxy for $d(\mathbf{X}, \mathbf{m})$. While this speeds up the
distance computation, the resulting estimates are now less reliable.
We therefore propose to interleave fast computations
$\hat{d}(\mathbf{X}, \mathbf{m})$ with the exact distance computations
$d(\mathbf{X}, \mathbf{m})$ by the following algorithm, which we refer
to as \AlgoNN{}:
\begin{algorithm}
\caption{\AlgoNN{}}
\label{alg:nn+gcc}
	\begin{algorithmic}[1]
		\State \textbf{Input:} $\mathbf{M}$, $\mathbf{X}$, $\hat{d}(\cdot, \cdot)$, $d(\cdot, \cdot)$, $\epsilon_S$
		\State $\mathbf{M}_{\text{done}} \leftarrow \emptyset$
    \For {$\mathbf{m} \in \mathbf{M}$} \Comment{First Loop}
      \State $\hat{\delta} \leftarrow \hat{d}(\mathbf{X}, \mathbf{m}$)
      \If {$\hat{\delta} > \epsilon_S$}
        \State $\delta \leftarrow d(\mathbf{X}, \mathbf{m}$)
        \State $\mathbf{M}_{\text{done}} \leftarrow \mathbf{M}_{\text{done}} \cup \mathbf{m}$
        \If {$\delta > \epsilon_S$}
          \State \textbf{return} $\mathbf{m}, \delta$
        \EndIf
      \EndIf
		\EndFor
    \State Sort $\mathbf{M} \setminus \mathbf{M}_{\text{done}}$ by $\hat{\delta}$ in descending order
    \For {$\mathbf{m} \in \mathbf{M} \setminus \mathbf{M}_{\text{done}}$} \Comment{Second Loop}
      \State $\delta \leftarrow d(\mathbf{X}, \mathbf{m})$
      \If {$\delta > \epsilon_S$}
        \State \textbf{return} $\mathbf{m}, \delta$
      \EndIf
    \EndFor
    \State \textbf{return} No collision-free motion found
	\end{algorithmic} 
\end{algorithm}

In a first loop, we run the network for each motion and only invoke
the GCC if a signed distance $\hat{\delta}$ greater than the safety
margin $\epsilon_S$ is predicted. If a collision-free motion is
found, the algorithm returns. In case no collision-free motion is
detected in the first loop, we resume with all motions which have not
yet been processed by the GCC. We sort the remaining motions according
to their predicted distances $\hat{\delta}$ in descending order. Then,
we run a second loop, where we check each motion with the GCC until a
collision-free motion is found or the algorithm returns with no
collision-free motion found. We thereby effectively reduce the
average number of exact collision checks that need to be executed
before finding the first collision-free motion in $\mathbf{M}$.

\subsection{Dataset Generation}
\label{sec:data_generation}

Our training datasets consist of tuples
$(\mathbf{x}, \mathbf{q}_0, \mathbf{q}_1, \delta)$ with query points
$\mathbf{x}$, the start and goal joint configurations, $\mathbf{q}_0$
and $\mathbf{q}_1$, and the label: the signed distance $\delta$.

Our method is applicable to any type of motion, which can be uniquely
defined by $\mathbf{q}_0$ and $\mathbf{q}_1$, since the exact motion
is implicitly defined via the training data. Therefore, we will
consider two motion types: a linear joint space motion, and a spline
motion in the task space (as used by KUKA Agilus robots). We will call
the respective datasets \DatasetLin{} and \DatasetSpline{}.

For a single motion, we first sample $\mathbf{q}_0$ and $\mathbf{q}_1$
uniformly from within the individual joint limits. For a linear joint
space motion, we discretize this motion into 200 steps by linearly
interpolating in joint space between $\mathbf{q}_0$ and
$\mathbf{q}_1$. For a spline motion, we use equidistant steps of the
flange along the spline in the task space. This yields a varying
number of steps per motion, though comparably many as for the linear
case. For a given discretized motion, we compute the sequence of poses
of each link of the robot. The associated geometries of a link along
with their pose sequence is the input for the swept volume
algorithm~\cite{sellan21tog} as implemented by the gpytoolbox
library~\cite{gpytoolbox}. This method outputs the swept volume as a
triangle mesh. We use an epsilon parameter of 15\,mm and otherwise use
the default parameters of the library. The swept volumes of the
individual link geometries are then merged into a final swept volume
by using the Boolean union operation of libigl~\cite{libigl}.

Note, that the discretization accuracy is thereby determined during
data generation, where arbitrary methods for swept volume generation
can be used. This only affects the quantity of training data and the
time which is necessary to train the models. The inference time of the
trained model is therefore mostly independent of the discretization
accuracy.

Given the mesh, we sample points in the task space and compute their
distances to the mesh. The points are sampled in a such a way, that we
have a global coverage of the task space but also pay attention to the
boundary of the mesh. For this, we first sample points uniformly in a
fixed bounding box large enough to cover any motion of the robot. We
use a density of 0.25 points per $\text{dm}^3$. Next we sample points
on the surface of the swept volume, using a density of
0.2
points per $\text{cm}^2$. For each such surface point we additionally
sample 2 points from a spherical Gaussian distribution centered at
this point (with $\sigma=3\,\text{mm}$). To further cover the space
inside the volume, we sample for each surface point a second surface
point at random, and define the point's location by uniformly sampling
from the line connecting these points.

\section{Experiments}

\begin{figure}[t]
  \centering
  \setlength\fboxsep{0pt}
  \hspace{0.95cm}
  \subfloat[ground truth]{\includegraphics[width=0.24\columnwidth]{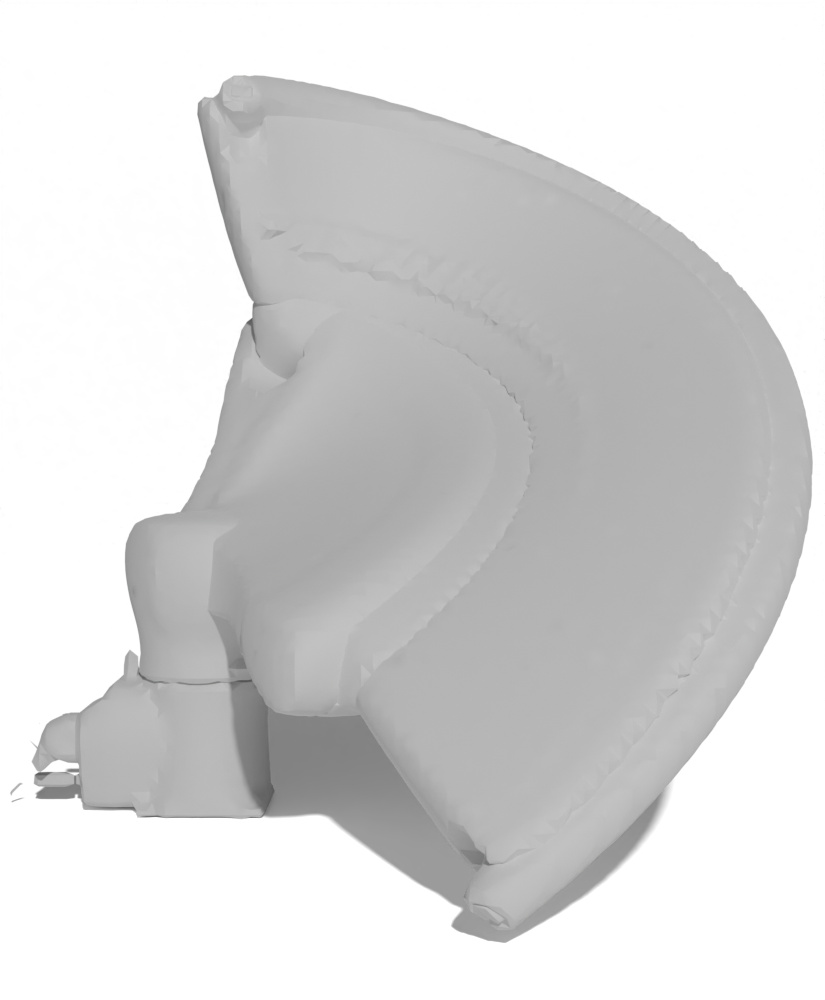}}
  \hfill
  \subfloat[11x1024]{
    \includegraphics[width=0.24\columnwidth]{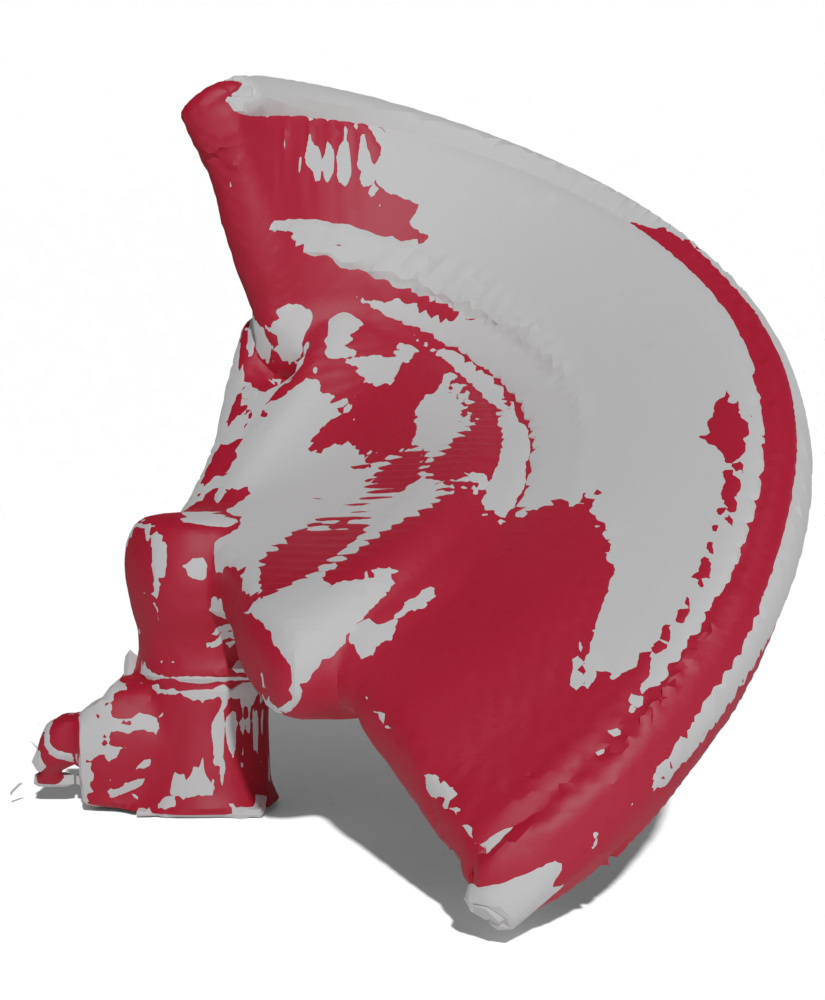}
    \hspace{-0.3cm}
    \includegraphics[width=0.24\columnwidth]{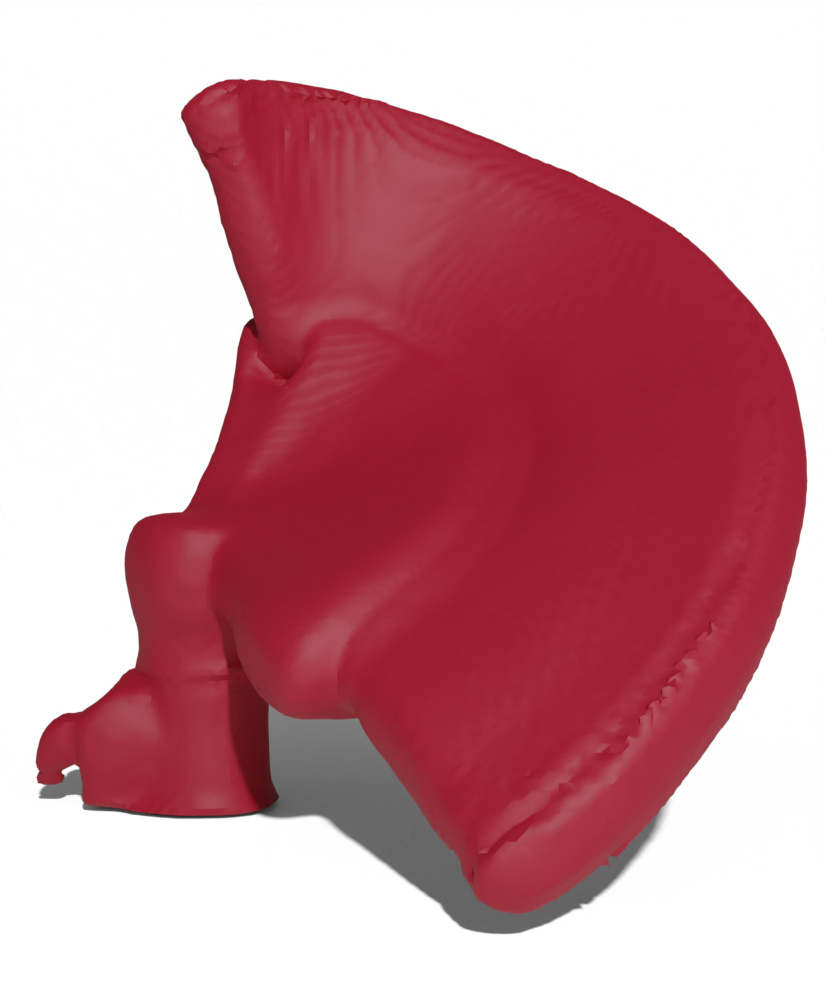}}

  \subfloat[11x512]{
    \includegraphics[width=0.24\columnwidth]{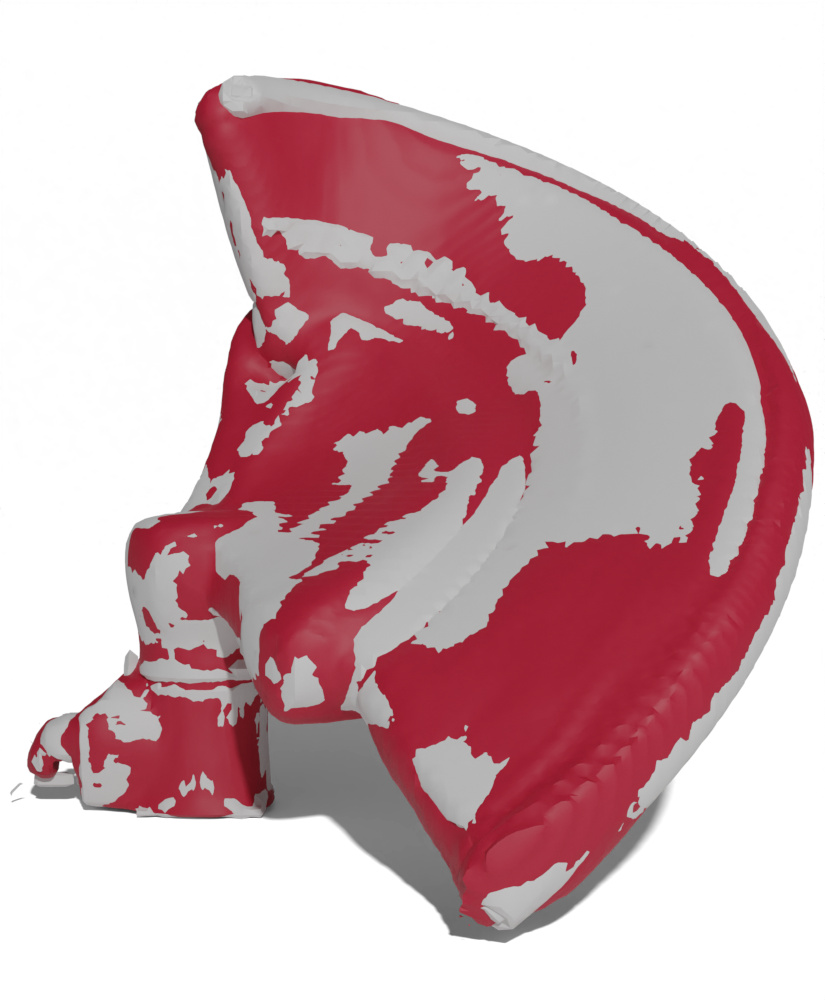}
    \hspace{-0.3cm}
    \includegraphics[width=0.24\columnwidth]{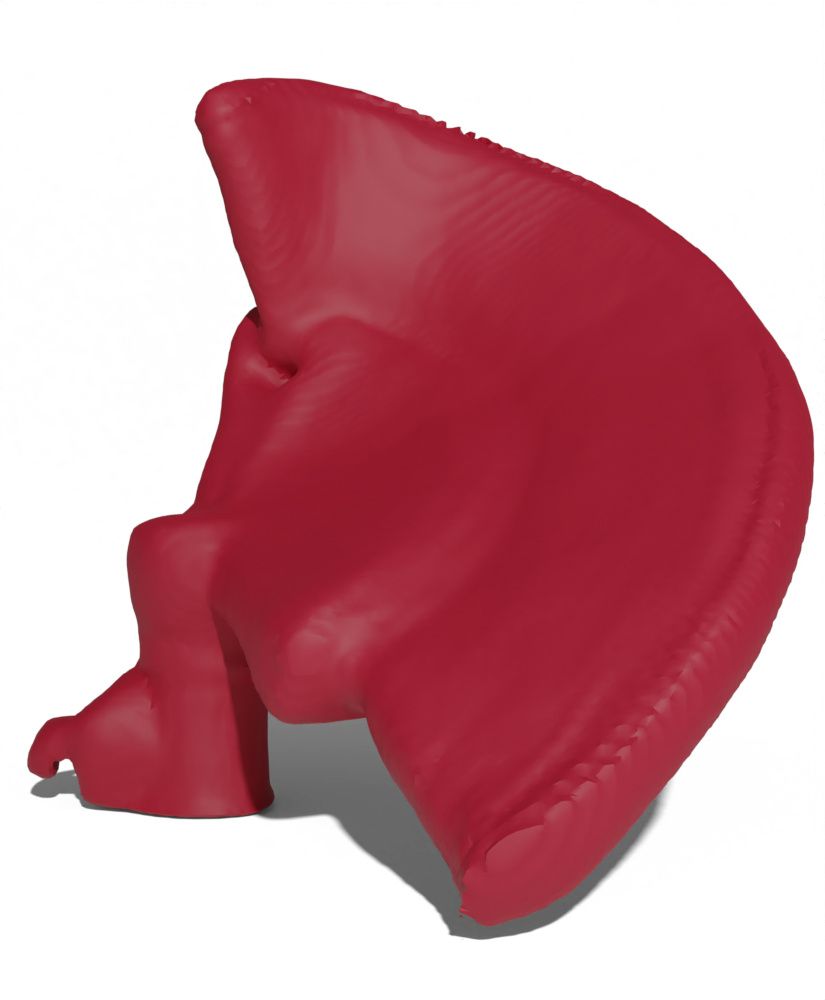}}
  \hfill
  \subfloat[5x512]{
    \includegraphics[width=0.24\columnwidth]{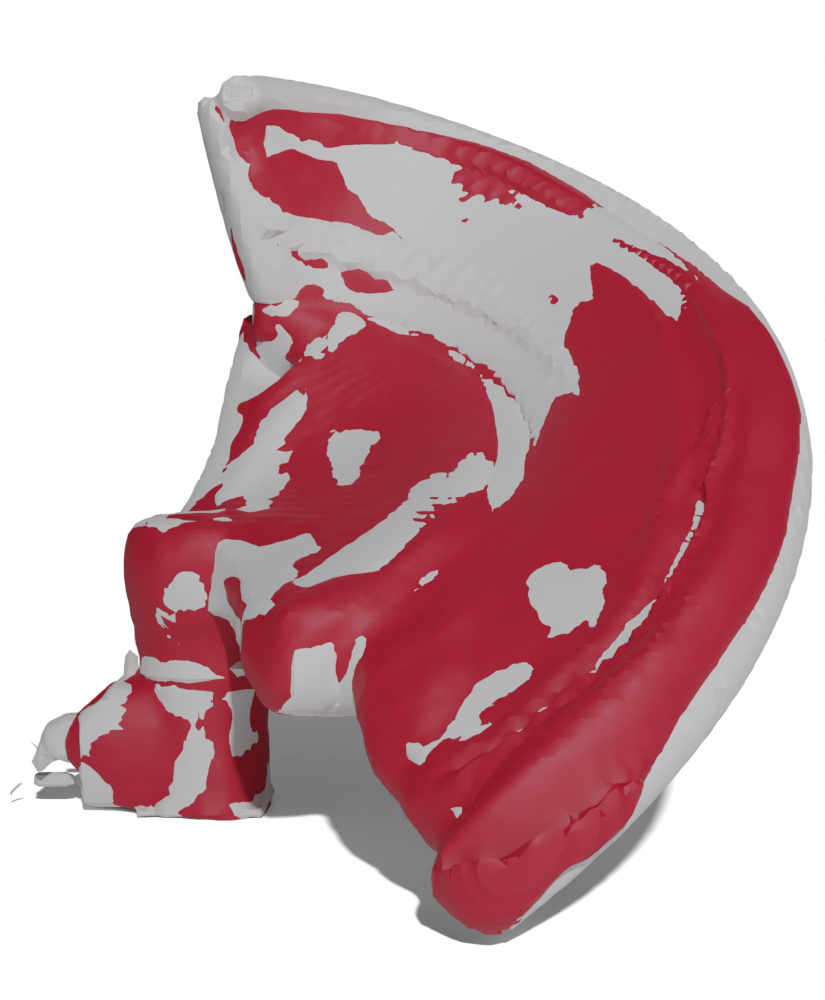}
    \hspace{-0.3cm}
    \includegraphics[width=0.24\columnwidth]{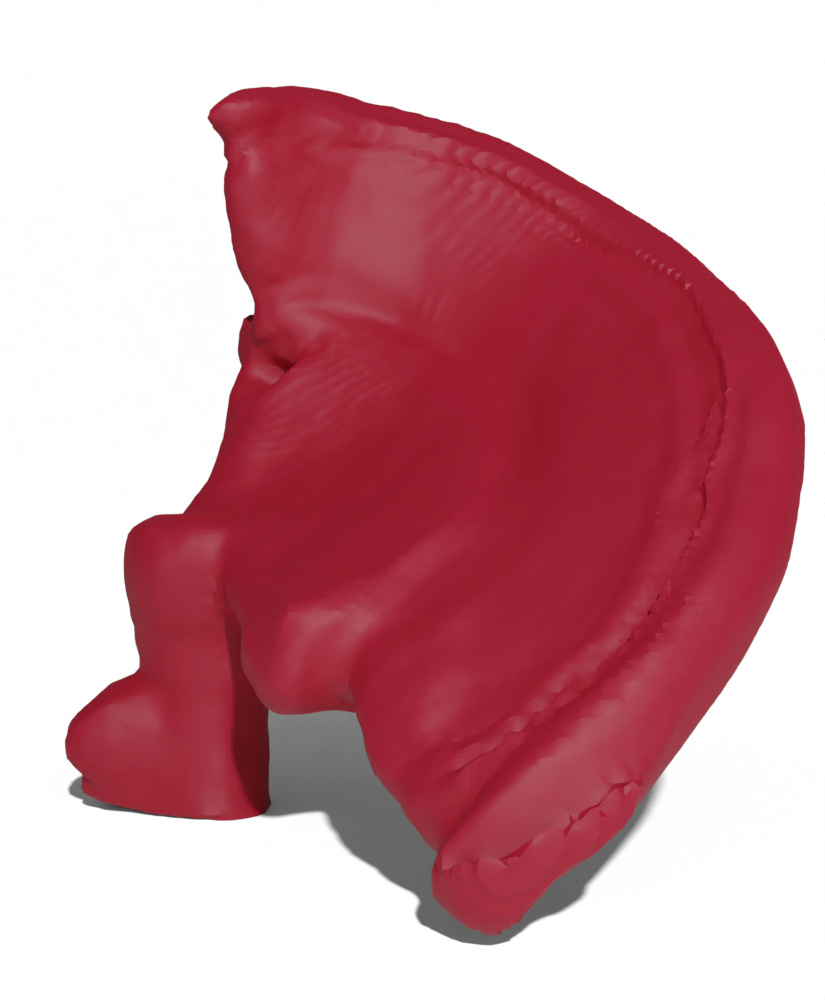}}
  \caption{Exemplary comparison of different model sizes for the
    linear joint space motion already shown in Fig.~\ref{fig:intro}.
    There is a noticeable degradation when going from (c)~11x512 to (d)~5x512,
    while the difference between (c)~11x512 and (b)~11x1024 is less
    pronounced. This qualitative finding is also reflected in the
    metrics shown in Table~\ref{tab:mae_lin_spline}.}
  \label{fig:modelSizeComp}
\end{figure}

\begin{figure*}[t]
  \vspace{2ex}
  \centering
  \includegraphics[width=0.075\textwidth]{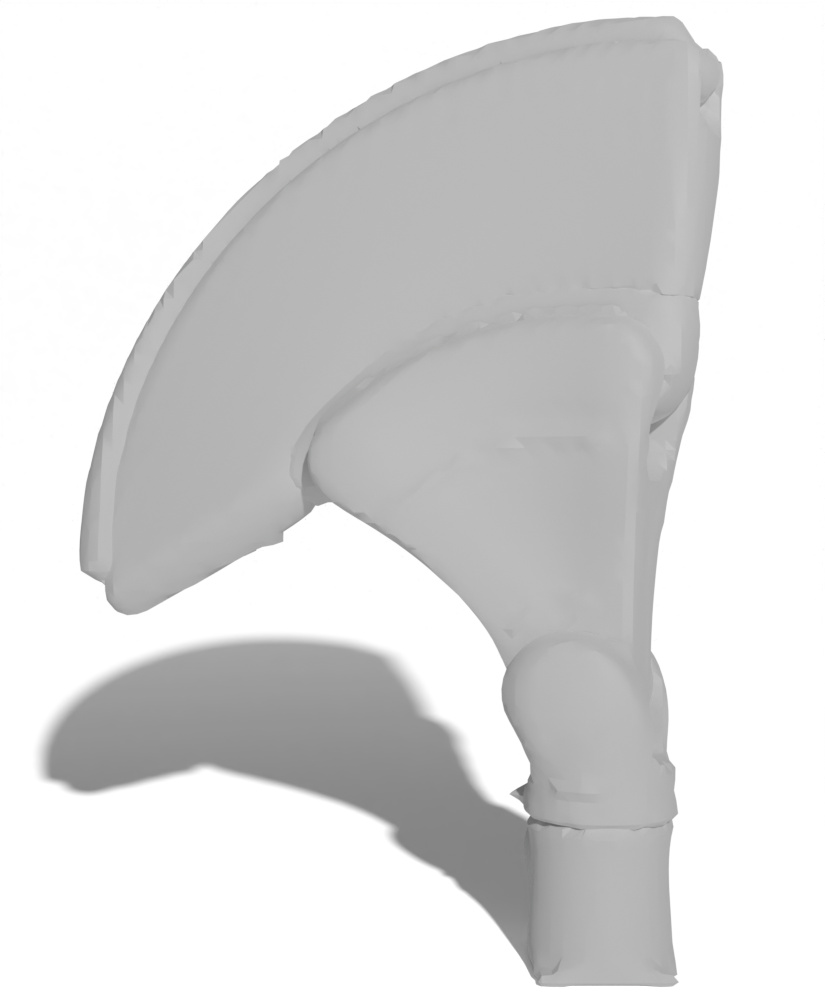}
  \includegraphics[width=0.075\textwidth]{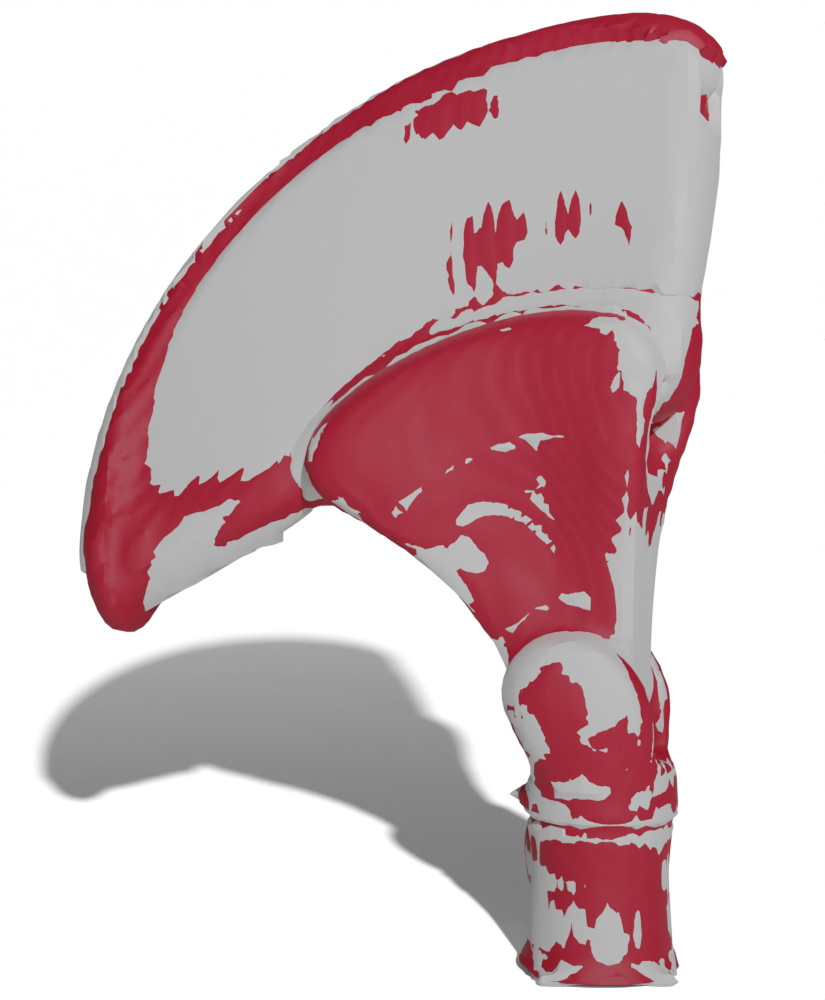}
  \includegraphics[width=0.075\textwidth]{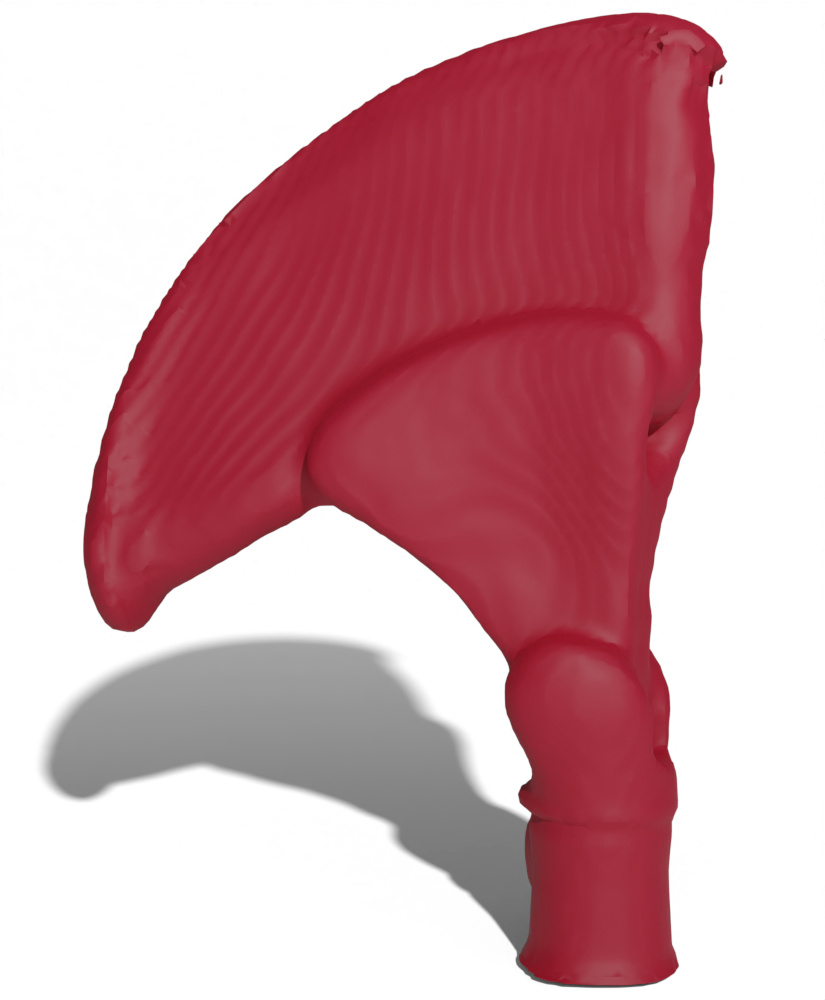}
  \hfill
  \includegraphics[width=0.075\textwidth]{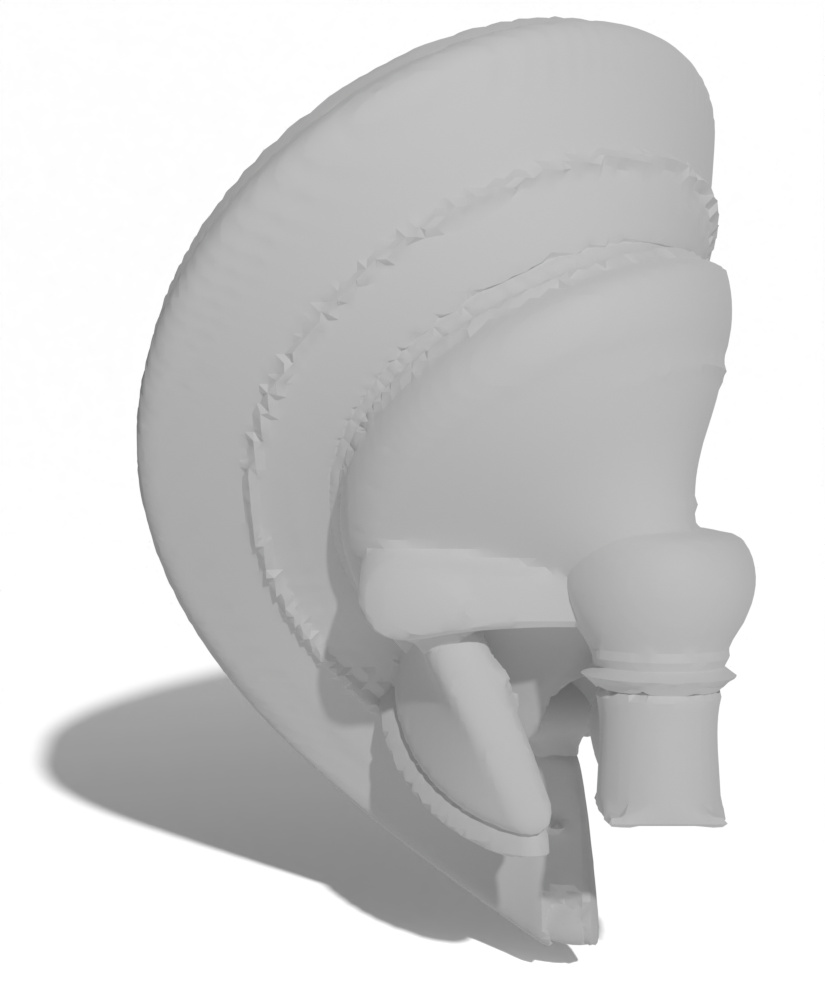}
  \includegraphics[width=0.075\textwidth]{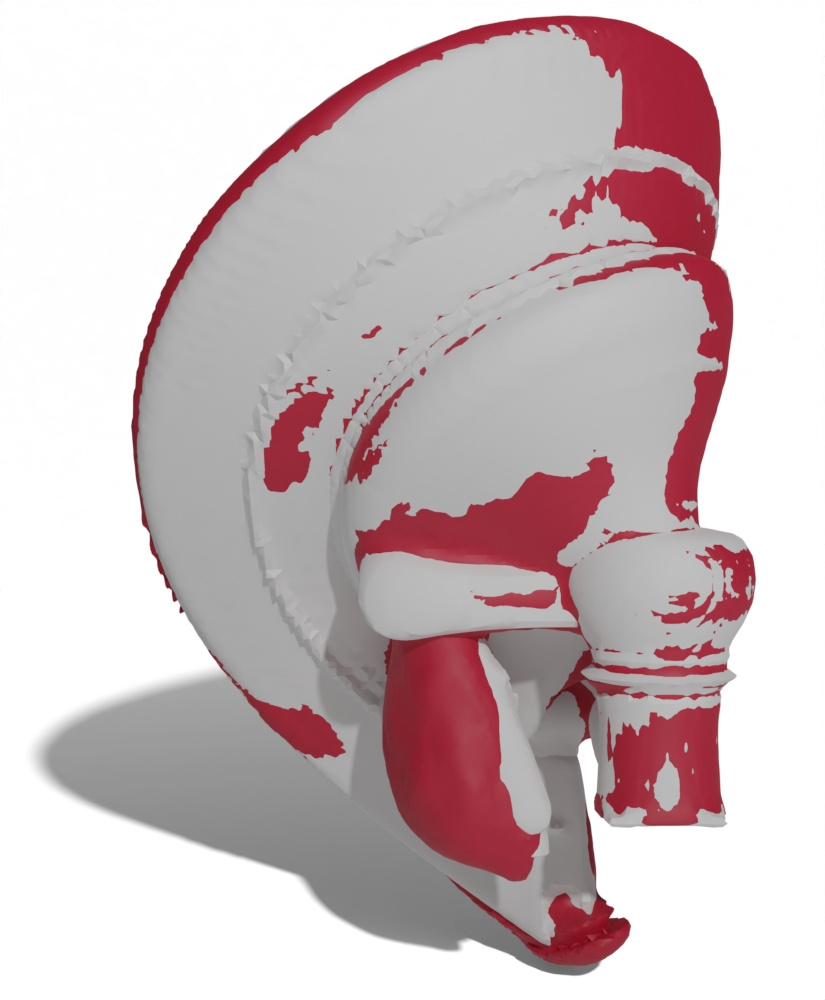}
  \includegraphics[width=0.075\textwidth]{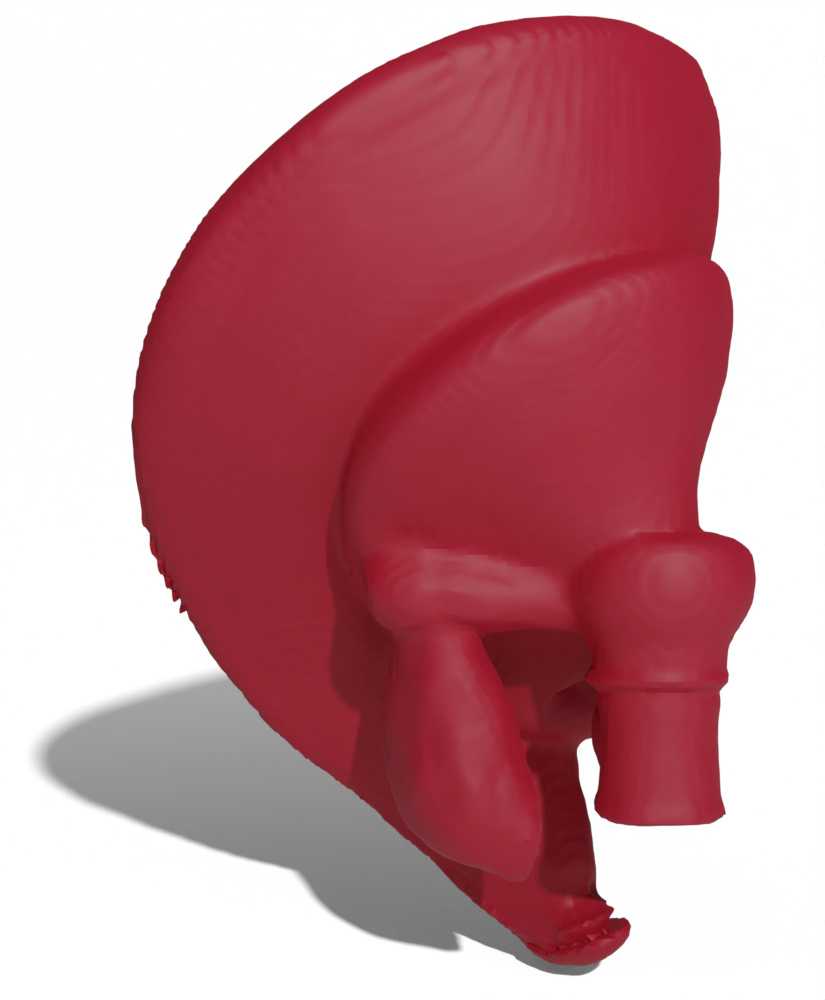}
  \hfill
  \includegraphics[width=0.075\textwidth]{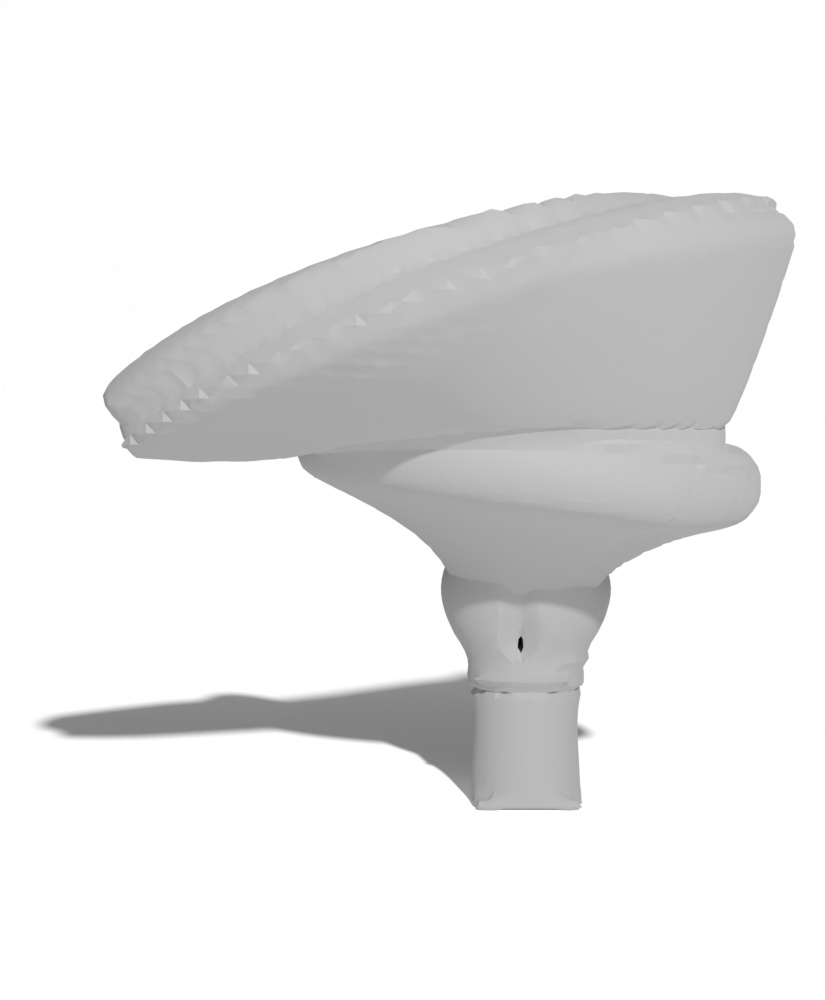}
  \includegraphics[width=0.075\textwidth]{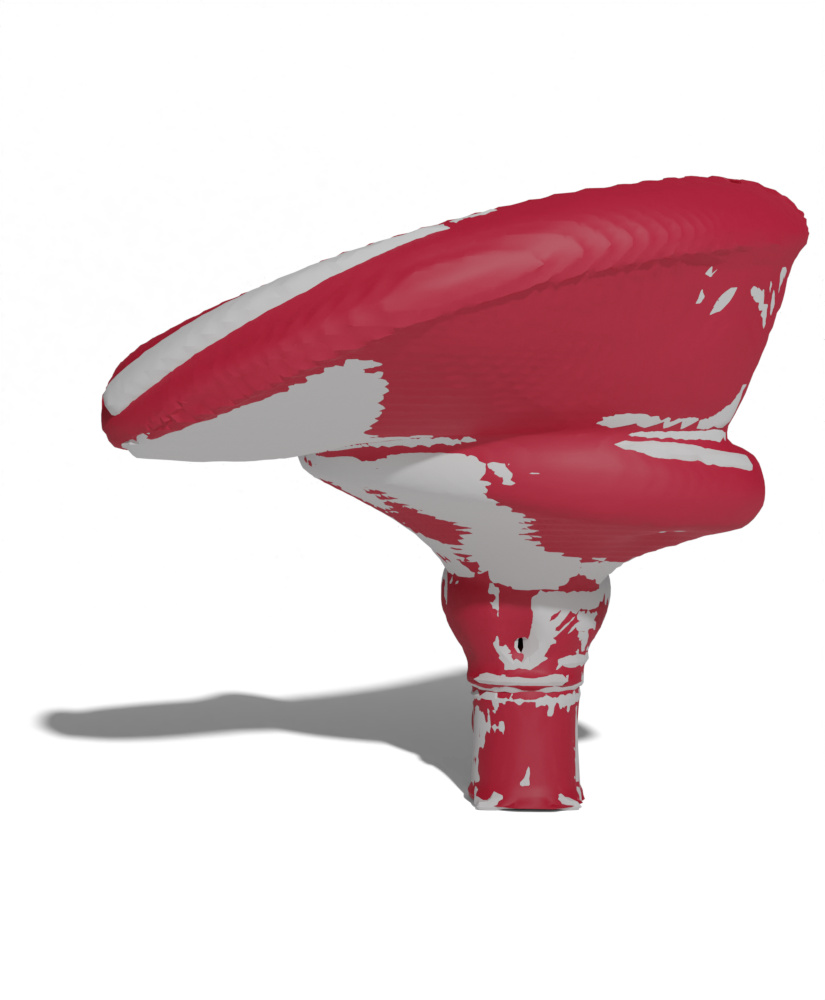}
  \includegraphics[width=0.075\textwidth]{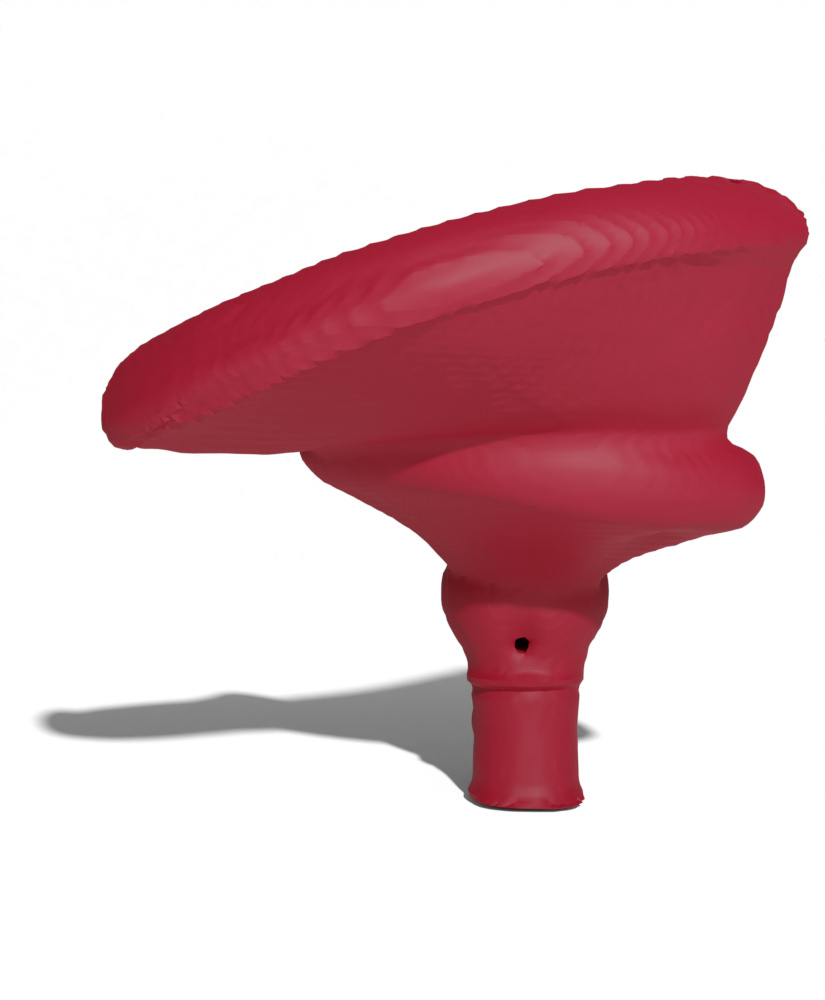}
  \hfill
  \includegraphics[width=0.075\textwidth]{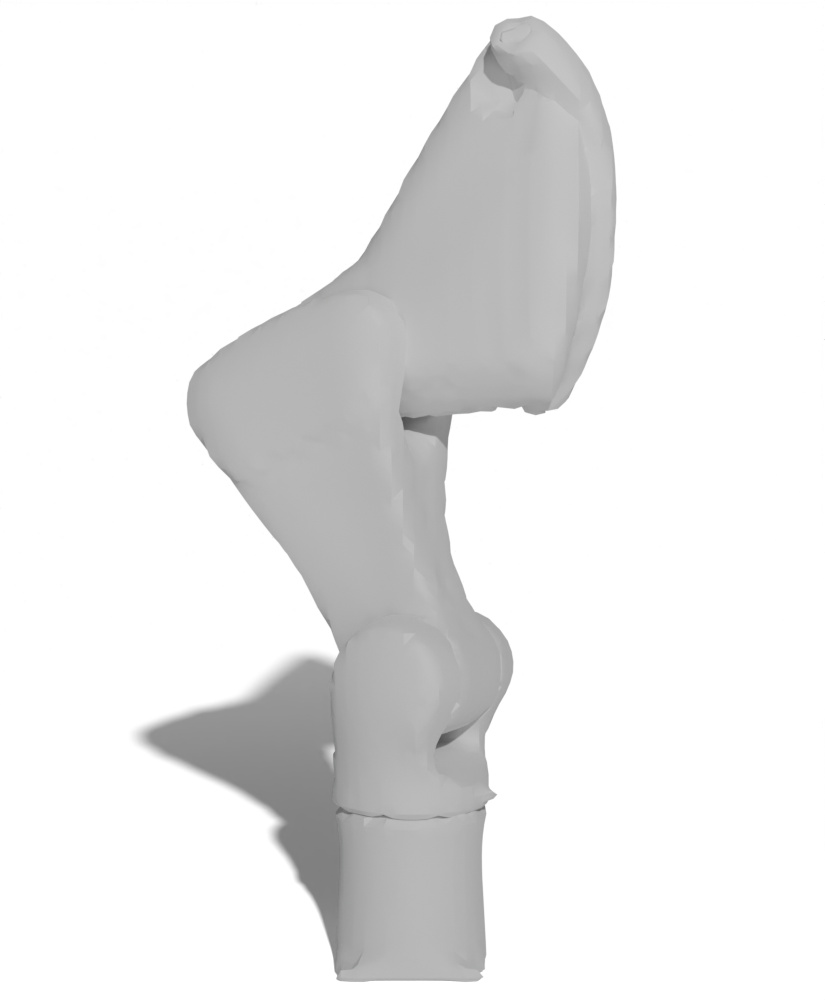}
  \includegraphics[width=0.075\textwidth]{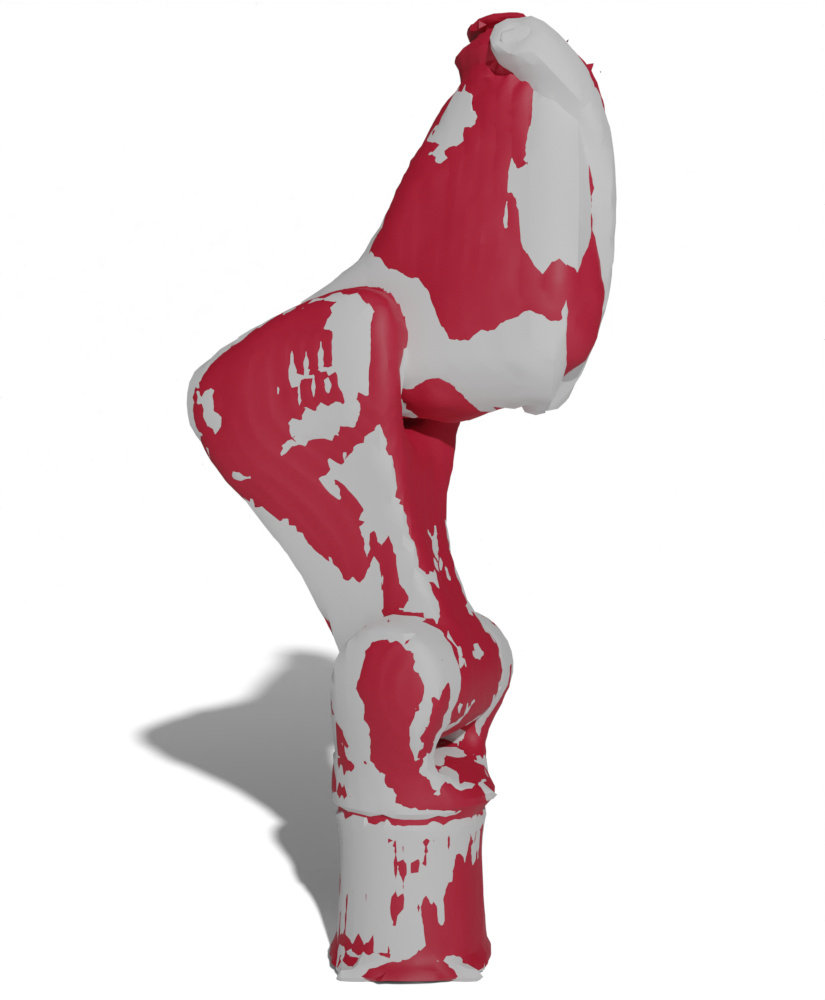}
  \includegraphics[width=0.075\textwidth]{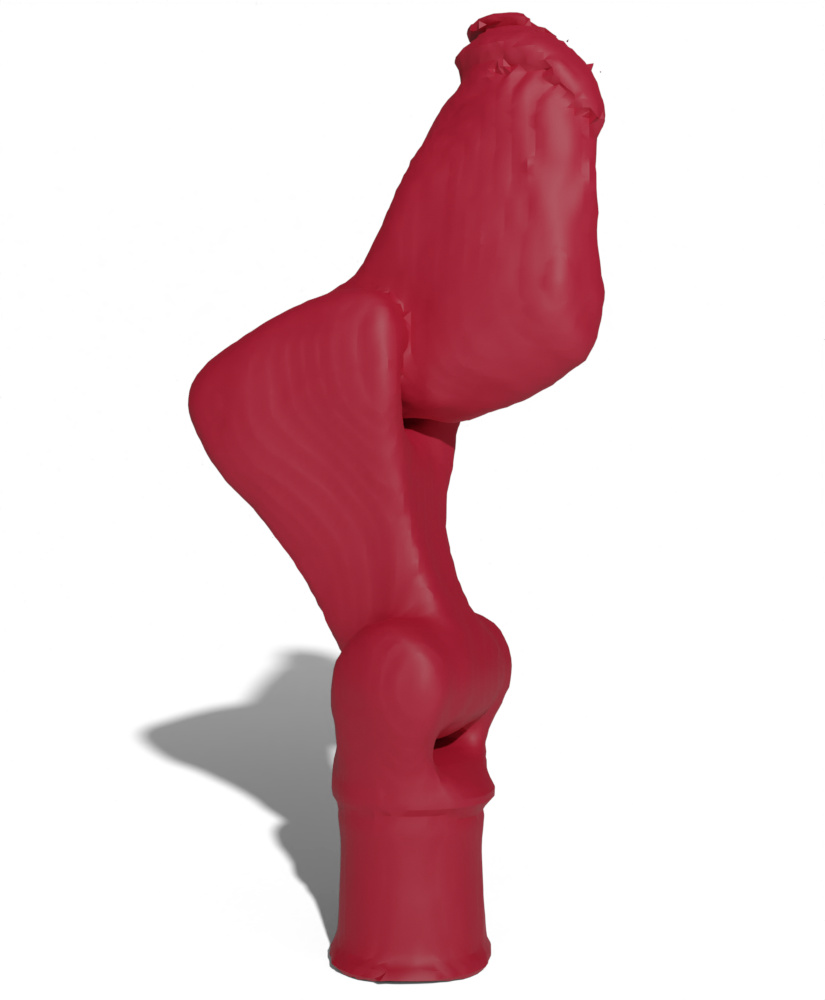}

  \includegraphics[width=0.075\textwidth]{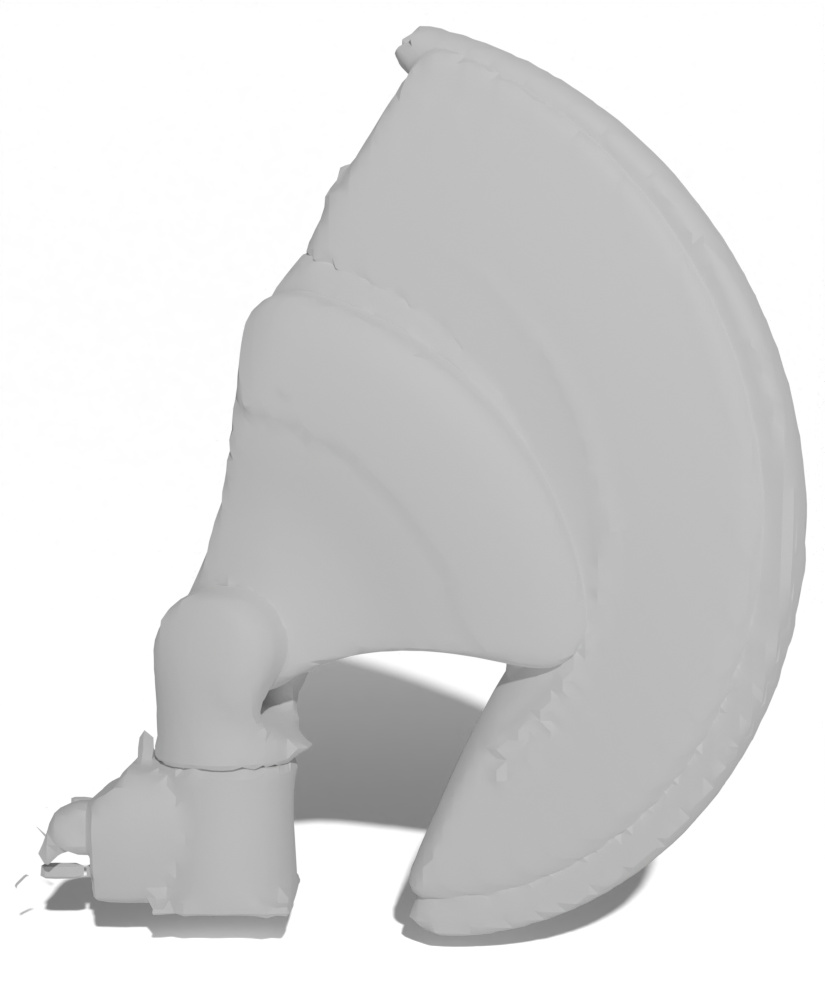}
  \includegraphics[width=0.075\textwidth]{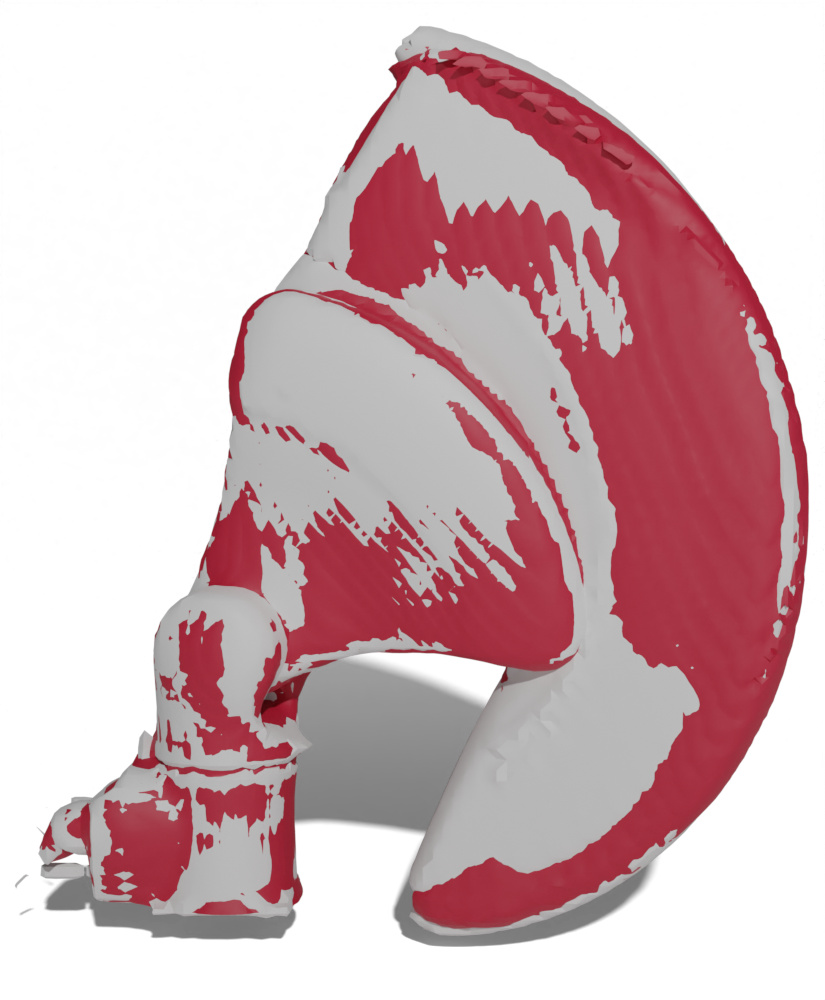}
  \includegraphics[width=0.075\textwidth]{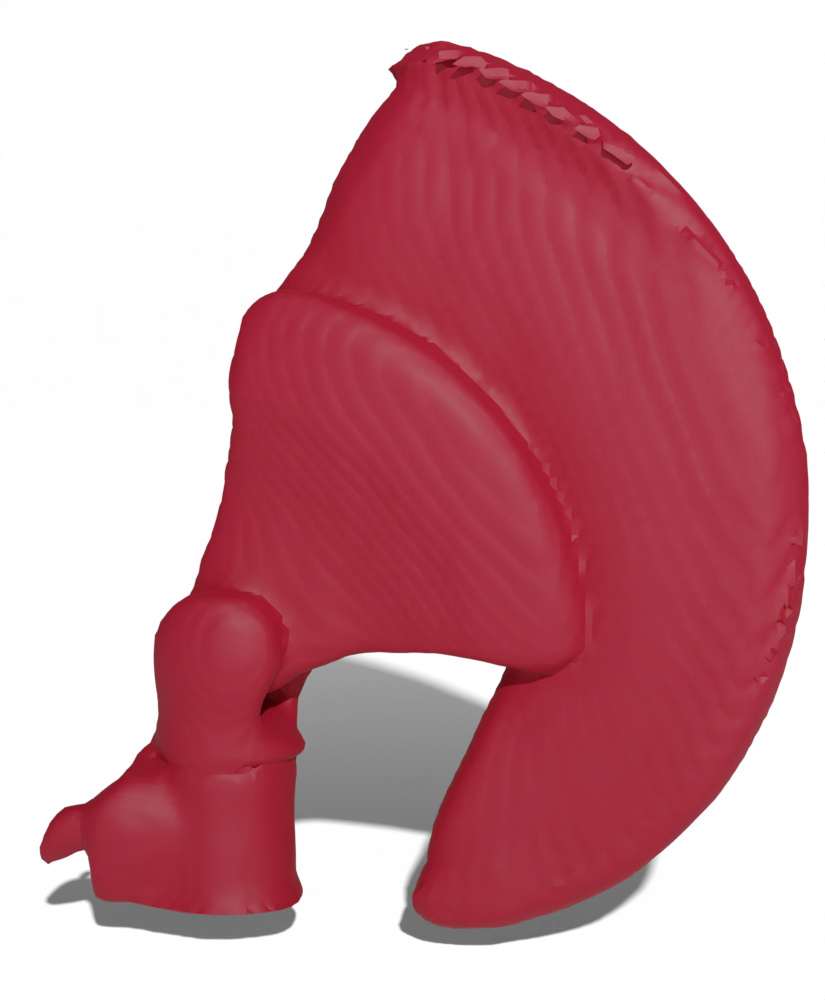}
  \hfill
  \includegraphics[width=0.075\textwidth]{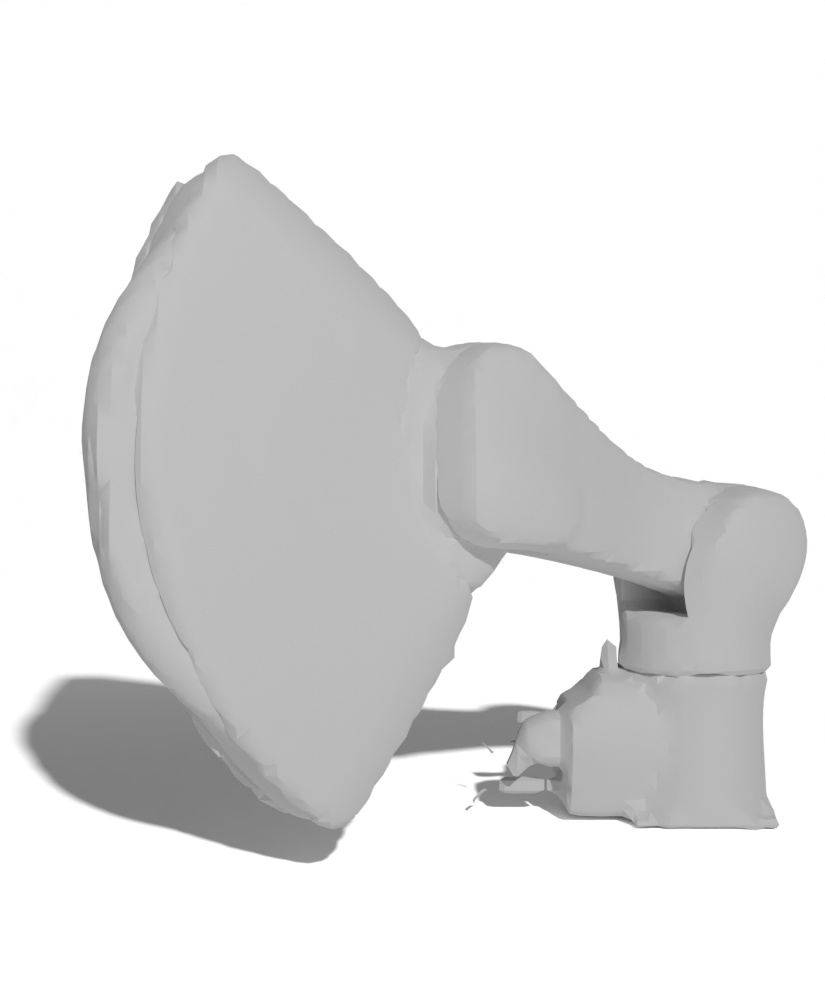}
  \includegraphics[width=0.075\textwidth]{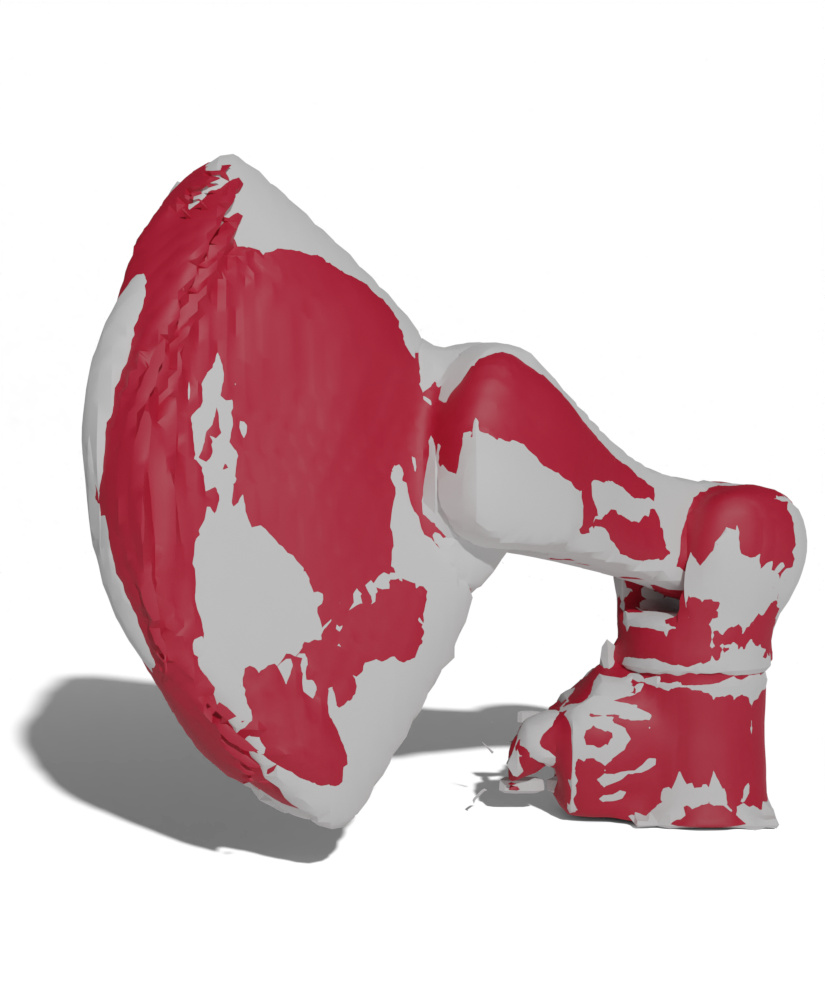}
  \includegraphics[width=0.075\textwidth]{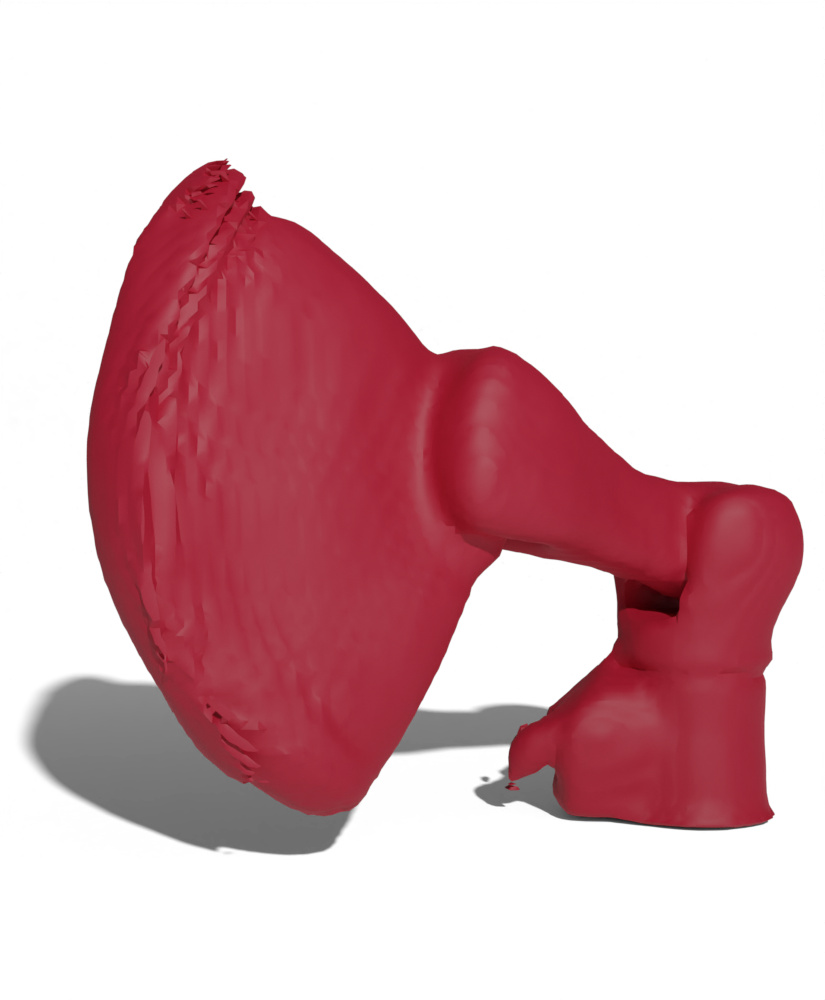}
  \hfill
  \includegraphics[width=0.075\textwidth]{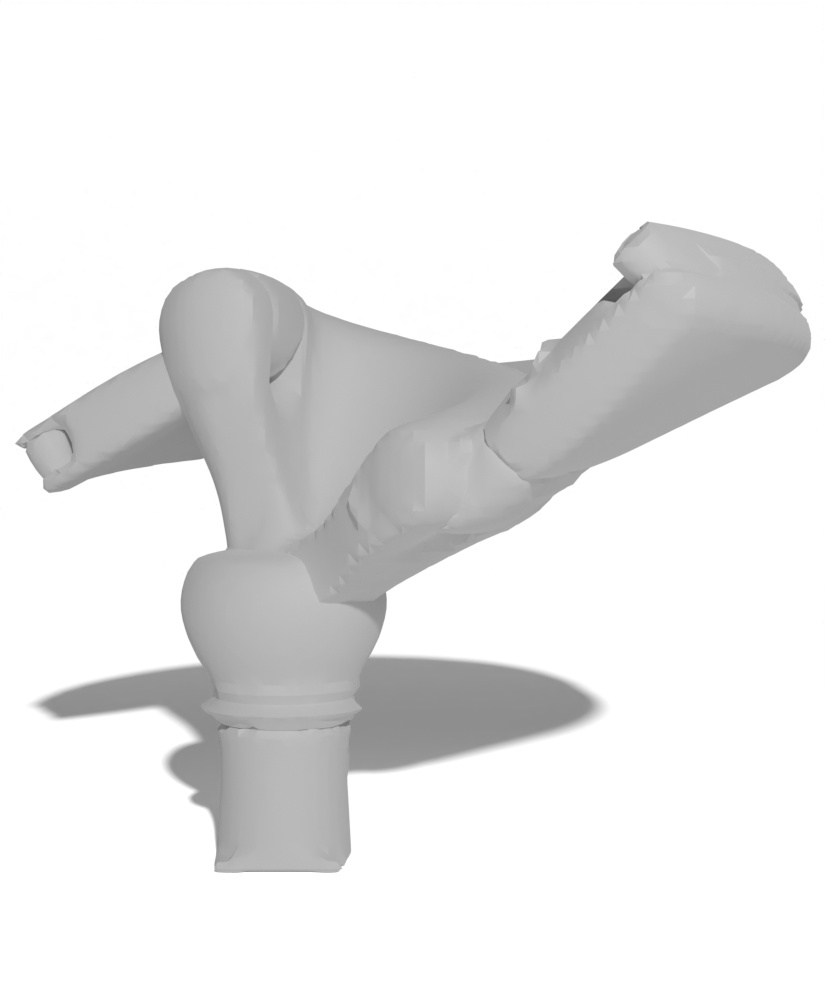}
  \includegraphics[width=0.075\textwidth]{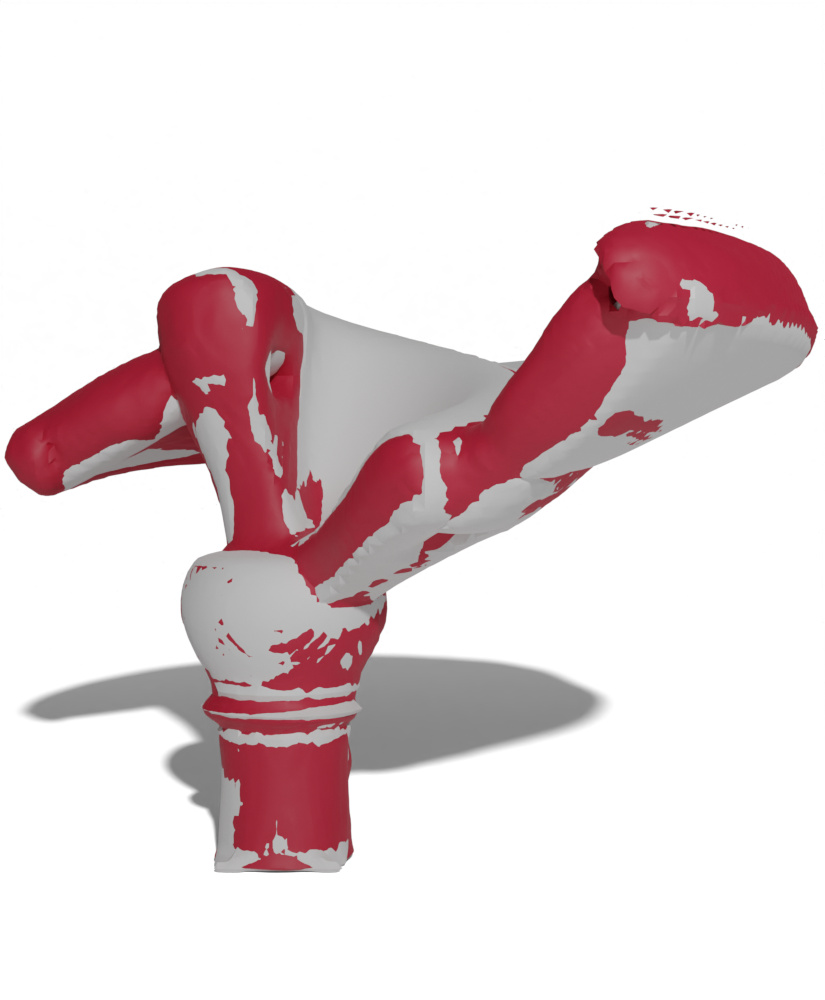}
  \includegraphics[width=0.075\textwidth]{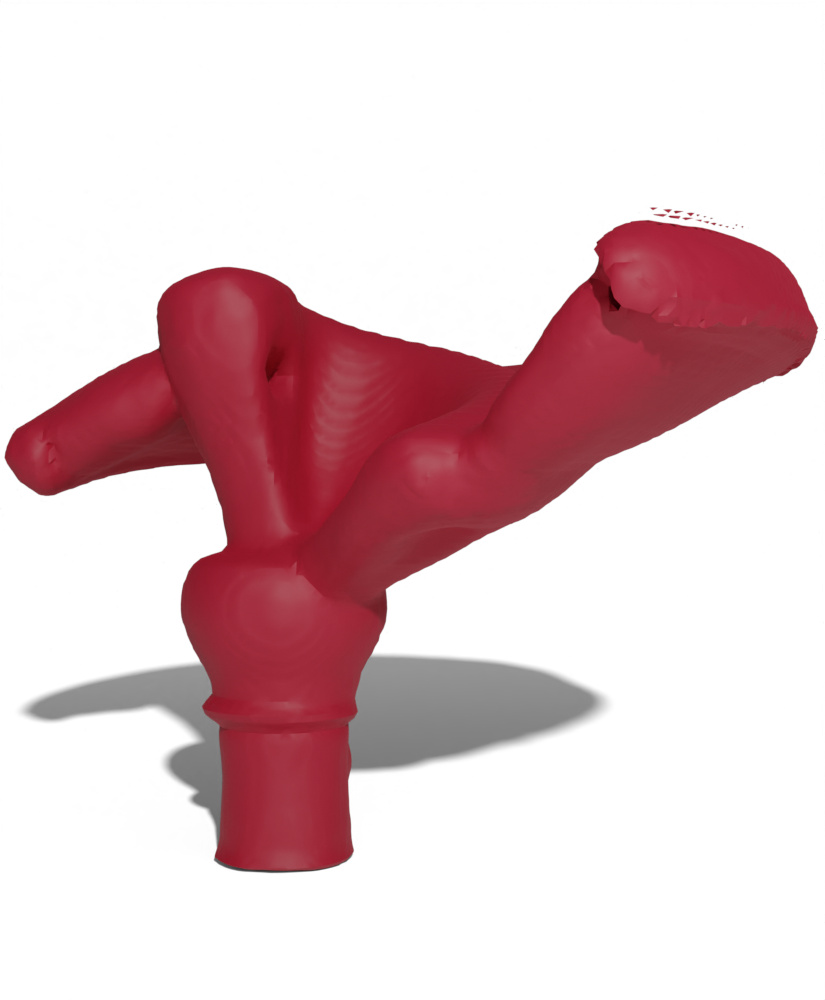}
  \hfill
  \includegraphics[width=0.075\textwidth]{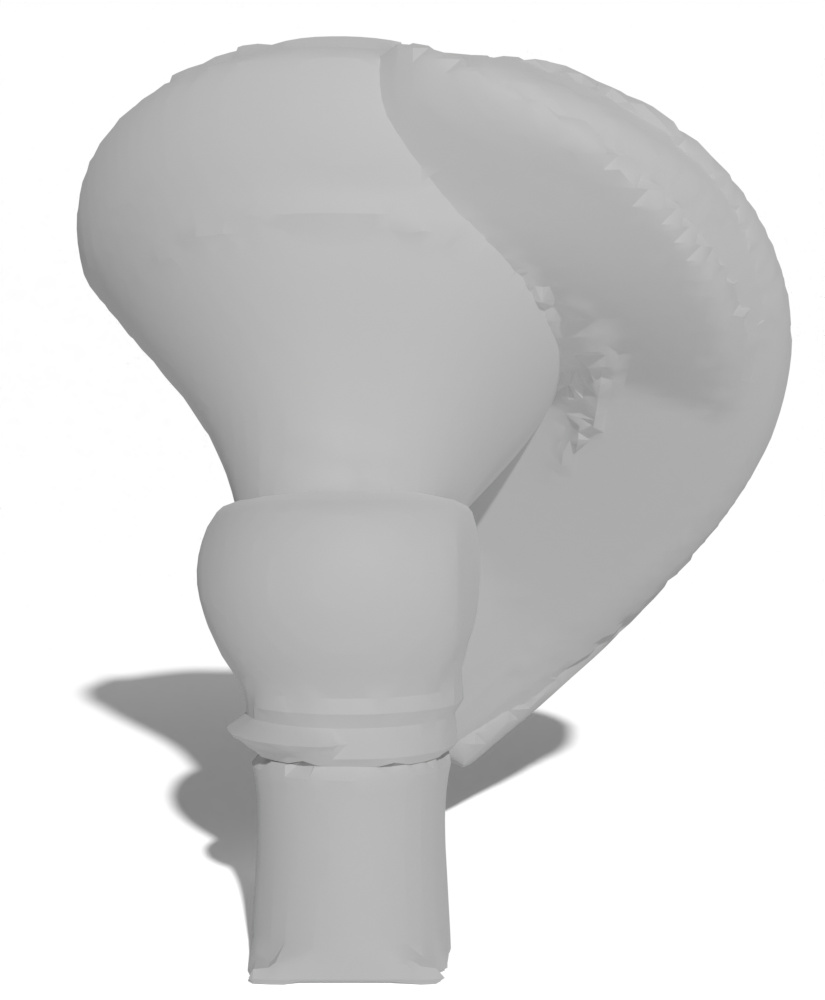}
  \includegraphics[width=0.075\textwidth]{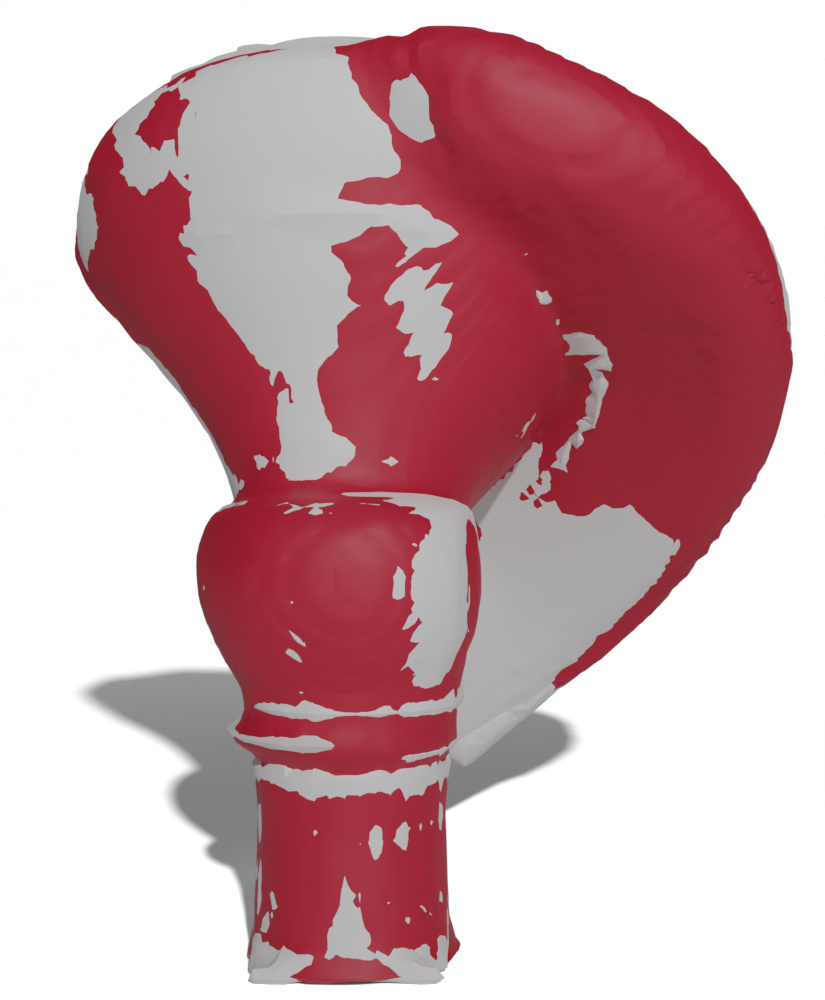}
  \includegraphics[width=0.075\textwidth]{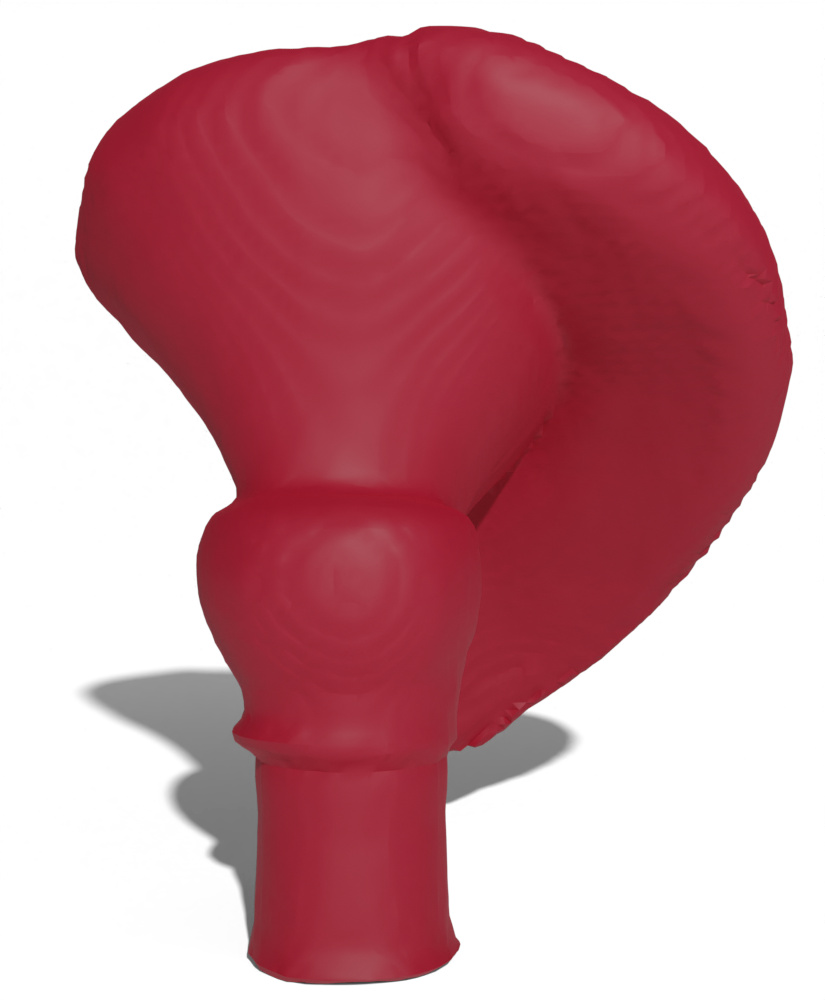}
  \caption{Comparison between the ground truth swept volume meshes
    (gray) and meshes reconstructed from the neural networks (red) via
    the marching cubes algorithm (resolution: 1\,cm). The upper row
    shows linear joint space motions, while the lower row shows spline
    motions. For both rows, 11x1024 models were used to predict the
    signed distances. The motions are the first four motions of their
    respective test sets, and thus have not been used in training.}
  \label{fig:testSetMotions}
  \vspace{-1ex}
\end{figure*}

We validate our approach in three different experiments. First, we
investigate the accuracy of the signed distance prediction of the
neural networks. We then evaluate Alg.~\ref{alg:nn+gcc} in an
artificial path planning scenario as well as a real-world bin picking
application.

If not otherwise stated, we use an \AgilusSmall{} with 6 degrees of
freedom. Computations are
done
with an Nvidia GeForce RTX 3090 GPU. We use the Flexible Collision
Library~(FCL)~\cite{fcl} as the GCC in all our experiments. For
training neural networks, we use the Adam optimizer of PyTorch with
default parameters lr~= 0.001 and betas~= (0.9, 0.999). We use the
ReduceLROnPlateau scheduler with factor 0.1 and patience 10 and train
the model for 200 epochs. A search over these parameters did not yield
any improvements. For testing, we choose the checkpoint with the best
loss on the validation set. Before inference, we export the PyTorch
models to TorchScript programs. Source code and checkpoints for the
models, as well as the dataset \DatasetLin{} are available
online~\cite{icra24sweptVolDataset}.

\subsection{Implicit Swept Volume Modeling}
\label{sec:sampling_motions}

In the first experiment, we test the accuracy of the signed distance
prediction on the two datasets \DatasetLin{} and \DatasetSpline{}
introduced in Section~\ref{sec:data_generation}. For each dataset we
generated 20\,000 motions and sampled approximately 4\,500 points per
motion. We use 60\,\% of the data for training, 25\,\% for
validation, and 15\,\% for testing. The mean absolute errors (MAE) on the test
sets are reported in Table~\ref{tab:mae_lin_spline}. We investigate
three different network architectures 5x512, 11x512, and 11x1024,
varying in their number of blocks $n_\text{b}$ (5 or 11) and their
block dimension $n_\text{dim}$ (512 or 1024). The MAE of the
prediction for both linear and spline motions is below 4.5\,mm in all
cases. For both datasets, we can improve the error to below
3.5\,mm by using 11 instead of 5 blocks and approximately 3.1\,mm by
increasing the block dimension from 512 to 1024.
To further illustrate the prediction quality of the networks, we used the
marching cubes algorithm to reconstruct the swept volume meshes from
the networks (with a resolution of 1\,cm).
Fig.~\ref{fig:modelSizeComp} shows a comparison of different model
sizes. More reconstructions for linear and spline motions are shown in
Fig.~\ref{fig:testSetMotions}.
\begin{table}[b]
  \small
  \centering
  \begin{tabular}{lll}
    \multicolumn{3}{l}{Mean Absolute Error (mm)}  \\\toprule
              & \DatasetLin{}  & \DatasetSpline{}      \\\midrule
      5x512   & 4.38 $\pm$5.67 & 4.32 $\pm$5.55   \\\midrule
      11x512  & 3.45 $\pm$4.71 & 3.32 $\pm$4.49   \\\midrule
      11x1024 & 3.08 $\pm$4.27 & 3.11 $\pm$4.28  \\\bottomrule
    \end{tabular}
    \caption{ Depicted are the mean absolute errors of the predicted
      signed distances. Results are provided for different model
      architectures and the two datasets.}
  \label{tab:mae_lin_spline}
\end{table}

We further report the inference times of our networks.
Table~\ref{tab:time_inference} shows average computation times for
processing three different sizes of point batches. The first two batch
sizes represent the point cloud sizes from the box world and bin
picking experiments, that we detail later. The inference times
approximately represent the time necessary to check one motion for
collision in these scenarios. We add a batch size of 20\,000 to
observe how inference times scale for larger inputs. We also trained a
16\,bit precision variant of the smallest model to further speed up
inference times. As our unit of measurement is millimeter, this has no
effect on the accuracy of the model and approximately halves the
computation time for larger batch sizes. For the two smaller batch
sizes of the path planning experiments, we observe that collision
checks can be performed in under a millisecond.

\begin{table}[t]
  \small
  \centering
  \begin{tabular}{lccl}
    \multicolumn{4}{l}{Inferences times (ms)}           \\\toprule
                  & Box World     & Bin Picking  &                \\\midrule
                  & 1000 Points  & 3000 Points & 20\,000 Points   \\\midrule
      11x1024     & 2.17   & 5.55 & 33.10 \\\midrule
      11x512      & 0.91   & 2.17 & 11.20 \\\midrule
      5x512       & 0.50   & 1.08 & 5.42  \\\midrule
      5x512 16bit & 0.49   & 0.70 & 2.63 \\\bottomrule
    \end{tabular}
  \caption{Inference times for different models and input sizes. 
  We report the time necessary to processes a batch of 1000, 3000 or 20\,000 points.
  The first two rows represent the point cloud sizes used in the path planning experiments Box World and Bin Picking.
  The standard deviation of the computation times is in the range of micro seconds and therefore omitted.}
  \label{tab:time_inference}
\end{table}

\subsection{Box World Collision Checking}

\begin{figure}[b] 
  \centering
  \includegraphics[width=0.6\columnwidth]{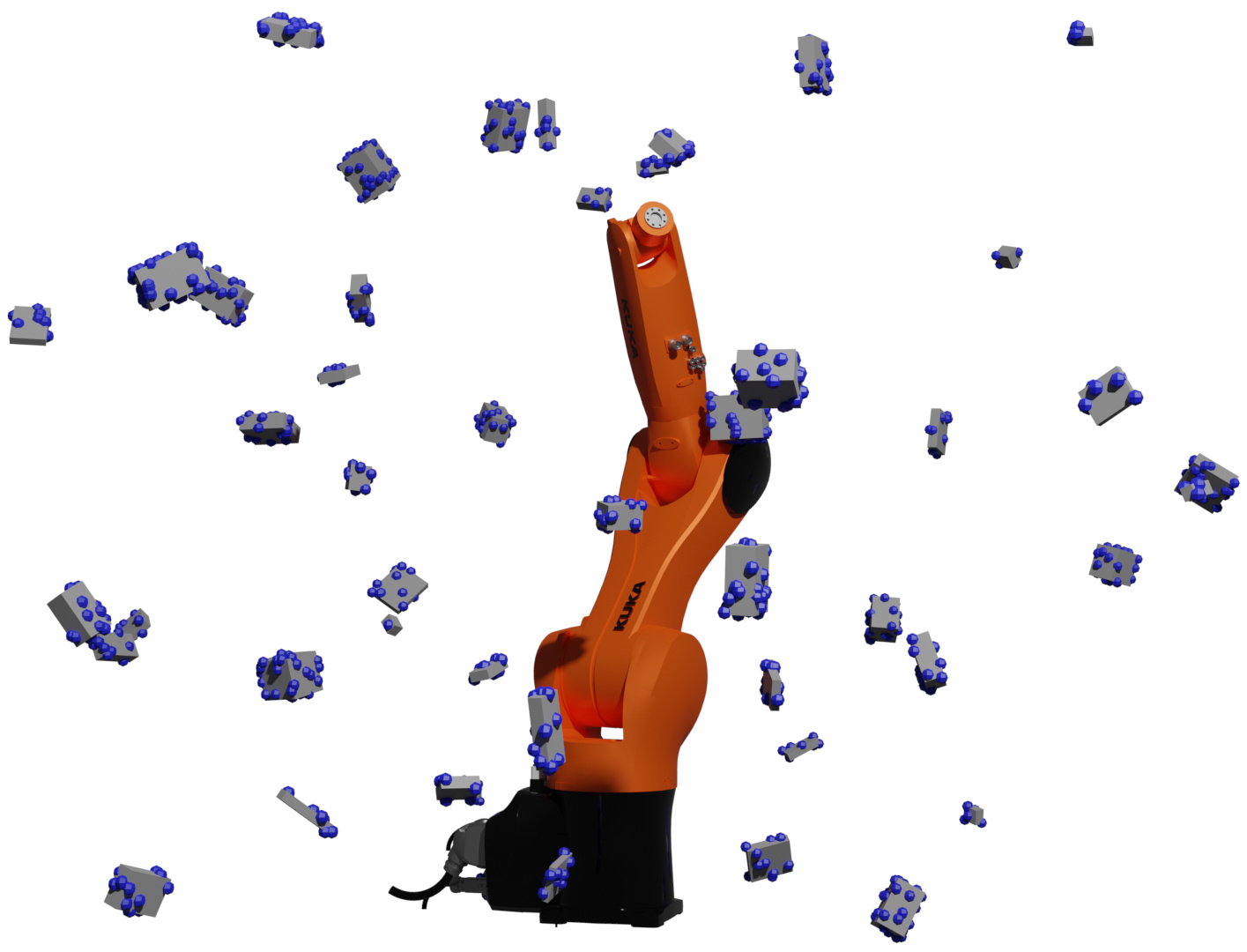}
  \caption{An example of a box world scene, with randomly sampled
    boxes and points on the surface of the boxes.}
  \label{fig:boxWorld}
\end{figure}

In the following experiment, we evaluate Alg.~\ref{alg:nn+gcc} for
motion planning in artificially generated box world scenes as seen in
Fig.~\ref{fig:boxWorld}.
Each of the 100 scenes contains 50 boxes with a random orientation and
a location sampled uniformly at random from within a volume defined by
$x,y \in [-1\,\text{m},\,1\,\text{m}]$ and
$z \in [0\,\text{m},\,1.5\,\text{m}]$. The side lengths of each box
are independently sampled uniformly from the interval
$[1\,\text{cm},\,10\,\text{cm}]$ (the robot base is centered at the
origin). For each scene, we sample a single start configuration
uniformly from within the joint limits, and 100 random goal
configurations, which yields 10\,000 motions in total. On average,
25.2\,\% of the motions are collision-free.
The minimum number of collision-free motions in a scene is 0, the
maximum is 66. For the neural network, we represent this environment
as a point cloud by sampling points from the surface of the boxes with
a density of 0.1 points per $\text{cm}^2$.
In the box world, the GCC reports the signed distance of the robot
model to the boxes represented as triangle meshes. For this, the
linear joint space motion is discretized into 200 steps and each
configuration is checked. We apply the \DatasetLin{} 5x512 16\,bit
model to predict signed distances. Using a safety margin of 5\,mm to
classify signed distances into collisions and non-collisions, the
network achieves a classification accuracy of 93.1\,\%, with 6.3\,\%
false positives, and 0.6\,\% false negatives.

We compare two algorithms for finding the first collision-free path of
each scene. The first method \AlgoGCC{} applies the GCC on each motion
in the given order. The second method \AlgoNN{} represents
Alg.~\ref{alg:nn+gcc} using the neural network as a proxy. As timing
depends on the order in which the motions of each scene are checked,
we average over 1000 permutations for each scene, which yields
100\,000 timing samples. The results shown in
Table~\ref{tab:box_world} indicate a speed-up of 49\,\% from 269\,ms
\AlgoGCC{} to 136\,ms \AlgoNN{} for the average time to find a
collision-free path. Due to the pre-checks done by the network, the
average number of exact collision checks done by the GCC is reduced to
less than a third, with 2.0 exact checks per scene in \AlgoNN{}.
\begin{table}[b]
  \small
  \centering
  \begin{tabular}{lll}
    \multicolumn{2}{c}{Results Box World Experiment}  \\\toprule
                      &          \AlgoGCC{}      & \AlgoNN{}            \\\midrule
      total time (ms) &  269 $\pm$286      & 136 $\pm$68        \\\midrule
      \# Exact checks &  6.69 $\pm$11.98   & 2.00 $\pm$9.85     \\\midrule
      \# NN checks    &                   & 14.05 $\pm$24.20   \\\bottomrule
    \end{tabular}
  \caption{Results of the box world experiment.
           \AlgoNN{} halves the total time for collision checking of \AlgoGCC{} by using three times less exact collision checks.}
  \label{tab:box_world}
\end{table}

\subsection{Bin Picking Collision Checking}

\begin{figure*}
\vspace{-3ex}
\end{figure*}

\begin{figure}[t]
  \centering
  \includegraphics[width=0.415\columnwidth]{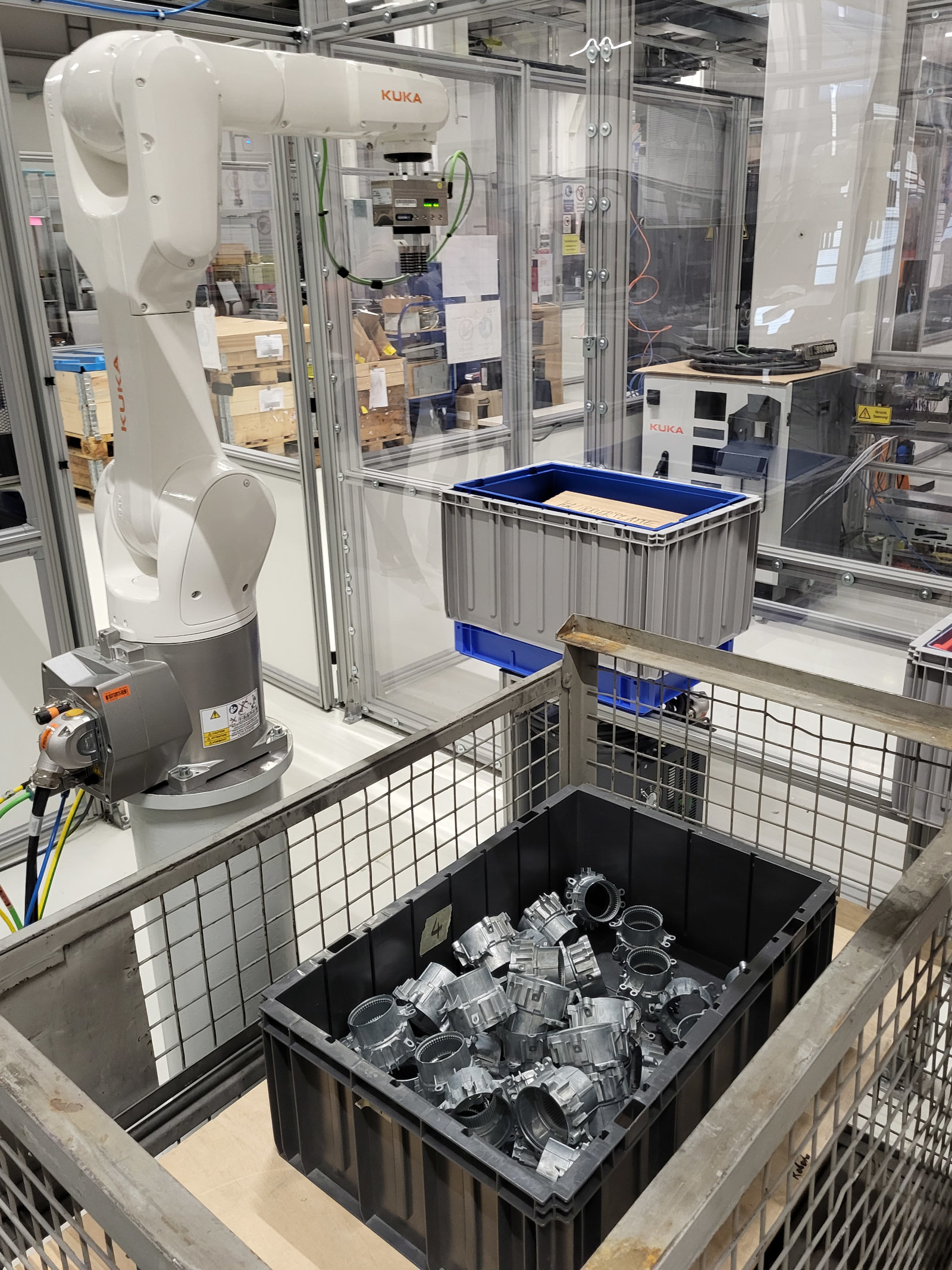}
  \includegraphics[width=0.535\columnwidth]{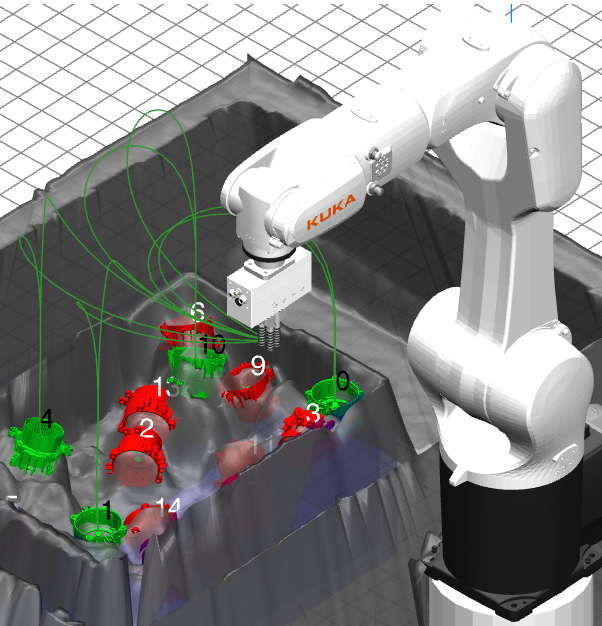}
  \caption{Bin picking application using the KUKA.SmartBinPicking
    software. The image on the left displays a test scenario with 40
    gearbox housings. The right image shows detected objects with
    collision-free endeffector paths (green) and objects without
    collision-free paths (red).}
  \label{fig:cell}
\end{figure}

Finally, we investigate collision checking in a real-world bin picking
application. Fig.~\ref{fig:cell} depicts a \AgilusBig{} (with attached
gripper) grasping gearbox housings out of a bin.

We use the KUKA.SmartBinPicking software to produce multiple grasp
candidates for a given scene. A given grasp \grasp{} is reached by two
splines \preStart{}~$\rightarrow$ \preVia{} and
\preVia{}~$\rightarrow$ \grasp{}, where \preStart{} is the starting
configuration above the bin, and \preVia{} is a via point located
close to the target object, which is introduced to increase the chance
for collision-free paths. Once the object has been grasped, it is
moved out of the bin using two similar splines: \grasp{}~$\rightarrow$
\postVia{} and \postVia{}~$\rightarrow$ \postEnd{}, where
\postEnd{}~$\neq$ \preStart{} is the final configuration above the
bin, and \postVia{}~$\neq$ \preVia{} is another via point. We refer to
the sequential execution of the first two splines as \motionPre{}, and
the other two splines as \motionPost{}. The GCC checks each of the
motions \motionPre{} and \motionPost{} in one pass, while the NN model
is called twice (once for each spline). For evaluating our approach,
we collected 22 real-world scenes (see Fig.~\ref{fig:cell}) with 15 to
40 gearbox housings in the bin. Grasp sampling of the system yielded
35~$\pm$7 possible grasp poses per scene where 82~$\pm$39\,\% of the
motions \motionPre{} to grasp candidates and 81~$\pm$39\,\%
of motions \motionPost{} out of the box resulted in collisions. 
A 5x512 16\,bit model is trained for the
new robot with attached gripper using the same data generation process
as described in Section~\ref{sec:sampling_motions}. While the network
was trained with the detailed model of the robot, the GCC uses a
convexified robot model for collision checking to speed up collision
checks.

Bin picking requires to precisely predict distances where the robot
geometry is close to the point cloud. This inherent difficulty
provides an interesting scenario, where the neural network is not able
to detect collision-free paths anymore for the given safety margin.
Alg.~\ref{alg:nn+gcc} is therefore reduced to the second loop: the
minimum signed distances of all possible motions are predicted by the
network in advance, and all motions are then checked by the GCC in
descending order of their predicted signed distance.

As can be seen, \AlgoNN{} can still improve the total time, if the
quality of the signed distance prediction is high enough to sort
motions favorably. Table~\ref{tab:bin_pick} shows that the number of
collision checks is reduced by approximately 50\,\% compared to \AlgoGCC{} for
motions into and out of the bin. 
The average time for checking a given motion is 28 $\pm4$\,ms for the
GCC and 0.70\,ms for the network. Thereby \AlgoNN{} achieves a
speed-up of 25\,\% and 30\,\% in total time consumption over the
\AlgoGCC{}. To produce the results, we again sampled 1000 permutations
for the order in which the motions are checked.

\begin{table}[t]
  \footnotesize
  \centering
  \begin{tabular}{lllll}
    & \multicolumn{2}{c}{\motionPre{} (into bin)} & \multicolumn{2}{c}{\motionPost{} (out of bin)}  \\\toprule
                      &    \AlgoGCC{}     & \AlgoNN{}        &    \AlgoGCC{}      & \AlgoNN{}           \\\midrule
      
      total time (ms) &  \scriptsize 190 $\pm$215     & \scriptsize 143 $\pm$188      &  \scriptsize 194 $\pm$223  & \scriptsize 135 $\pm$185     \\\midrule
      \# Exact checks &  \scriptsize 6.90 $\pm$8.30 & \scriptsize 3.38 $\pm$7.32  &  \scriptsize 7.2 $\pm$8.7  & \scriptsize 3.25 $\pm$7.3        \\\midrule
      \# NN checks    &                             & \scriptsize 70.5 $\pm$13.00 &  \scriptsize               & \scriptsize 70.5 $\pm$13.0      \\\bottomrule
    \end{tabular}\scriptsize 
  \caption{Results of the bin picking experiment:
           Performance of \AlgoGCC{} and \AlgoNN{} is shown for motions into the bin in columns 2 and 3 
           and motions out of the bin in columns 4 and 5.
           It can be seen, that \AlgoNN{} improves the total time for collision checking by 25\,\% and 30\,\%, respectively,
           using less exact collision checks than \AlgoGCC{}.}
  \label{tab:bin_pick}
\end{table}

\section{Conclusion}

We proposed a novel approach for neural implicit swept volumes,
parameterized by the start and goal configuration. We further
introduced an algorithm for interleaving these implicit models with
geometric collision checking. In our experiments, we achieve accurate
neural representations of swept volumes and show, that we can
significantly speed up a real-world bin picking application with our
algorithm.

\section*{Acknowledgments}

This work has been partially supported by the European Commission (EC)
under grant agreement ID 101091792 (SMARTHANDLE) and by the Bavarian
Ministry of Economic Affairs, Regional Development and Energy (StMWi)
under contract number DIK0374/01 (OPERA).

\bibliographystyle{IEEEtran}
\bibliography{strings, swept_vol}

\begin{thebibliography}{10}
\providecommand{\url}[1]{#1}
\csname url@rmstyle\endcsname
\providecommand{\newblock}{\relax}
\providecommand{\bibinfo}[2]{#2}
\providecommand\BIBentrySTDinterwordspacing{\spaceskip=0pt\relax}
\providecommand\BIBentryALTinterwordstretchfactor{4}
\providecommand\BIBentryALTinterwordspacing{\spaceskip=\fontdimen2\font plus
\BIBentryALTinterwordstretchfactor\fontdimen3\font minus
  \fontdimen4\font\relax}
\providecommand\BIBforeignlanguage[2]{{%
\expandafter\ifx\csname l@#1\endcsname\relax
\typeout{** WARNING: IEEEtran.bst: No hyphenation pattern has been}%
\typeout{** loaded for the language `#1'. Using the pattern for}%
\typeout{** the default language instead.}%
\else
\language=\csname l@#1\endcsname
\fi
#2}}

\bibitem{kleeberger2020crr}
K.~Kleeberger, R.~Bormann, W.~Kraus, and M.~F. Huber,
  ``\href{https://doi.org/10.1007/s43154-020-00021-6}{A Survey on
  Learning-Based Robotic Grasping},'' \emph{Current Robotics Reports}, vol.~1,
  pp. 239--249, 12 2020.

\bibitem{tenpas2017ijrr}
A.~ten Pas, M.~Gualtieri, K.~Saenko, and R.~Platt~Jr.,
  ``\href{https://doi.org/10.1177/0278364917735594}{Grasp Pose Detection in
  Point Clouds},'' \emph{Int. Journal of Robotics Research ({IJRR})}, vol.~36,
  no. 13--14, pp. 1455--1473, 2017.

\bibitem{pan15engineering}
J.~Pan and D.~Manocha,
  ``\href{https://doi.org/10.15302/J-ENG-2015009}{Efficient Configuration Space
  Construction and Optimization for Motion Planning},'' \emph{Engineering},
  vol.~1, no.~1, pp. 46--57, Mar. 2015.

\bibitem{coleman14joser}
D.~Coleman, I.~A. Șucan, S.~Chitta, and N.~Correll,
  ``\href{https://doi.org/10.6092/JOSER_2014_05_01_p3}{Reducing the Barrier to
  Entry of Complex Robotic Software: a MoveIt! Case Study},'' \emph{Journal of
  Software Engineering for Robotics (JOSER)}, vol.~5, no.~1, pp. 3--16, Mar.
  2014.

\bibitem{sellan21tog}
S.~Sell\'{a}n, N.~Aigerman, and A.~Jacobson,
  ``\href{https://doi.org/10.1145/3450626.3459780}{Swept Volumes via Spacetime
  Numerical Continuation},'' \emph{{ACM} Trans. on Graphics}, vol.~40, no.~4,
  pp. 1--11, 2021.

\bibitem{kaidi2019neurips}
K.~Cao, C.~Wei, A.~Gaidon, N.~Arechiga, and T.~Ma,
  ``\href{https://proceedings.neurips.cc/paper_files/paper/2019/file/621461af90cadfdaf0e8d4cc25129f91-Paper.pdf}{Learning
  Imbalanced Datasets with Label-Distribution-Aware Margin Loss},'' in
  \emph{Advances in Neural Information Processing Systems ({NeurIPS})}, 2019.

\bibitem{tan2020cvpr}
J.~Tan, C.~Wang, B.~Li, Q.~Li, W.~Ouyang, C.~Yin, and J.~Yan,
  ``\href{https://openaccess.thecvf.com/content_CVPR_2020/papers/Tan_Equalization_Loss_for_Long-Tailed_Object_Recognition_CVPR_2020_paper.pdf}{Equalization
  Loss for Long-Tailed Object Recognition},'' in \emph{Proc.~of the {IEEE}
  Conf.~on Computer Vision and Pattern Recognition {(CVPR)}}, 2020.

\bibitem{wang2020iclr}
X.~Wang, L.~Lian, Z.~Miao, Z.~Liu, and S.~Yu,
  ``\href{https://openreview.net/pdf?id=D9I3drBz4UC}{Long-tailed Recognition by
  Routing Diverse Distribution-Aware Experts},'' in \emph{Proc.~of the
  Int.~Conf.~on Learning Representations (ICLR)}, 2021.

\bibitem{chiang20wafr}
H.-T.~L. Chiang, A.~Faust, S.~Sugaya, and L.~Tapia,
  ``\href{https://doi.org/10.1007/978-3-030-44051-0_4}{Fast Swept Volume
  Estimation with Deep Learning},'' in \emph{Algorithmic Foundations of
  Robotics XIII: Proc. of the 13th Workshop on the Algorithmic Foundations of
  Robotics {(WAFR)}}, 2020, {DOI}: 10.1007/978-3-030-44051-0\_4.

\bibitem{baxter20iros}
J.~Baxter, M.~R. Yousefi, S.~Sugaya, M.~Morales, and L.~Tapia,
  ``\href{https://doi.org/10.1109/IROS45743.2020.9341396}{Deep Prediction of
  Swept Volume Geometries: Robots and Resolutions},'' in \emph{Proc.~of the
  {IEEE}/{RSJ} Int.~Conf.~on Intelligent Robots and Systems ({IROS})}, 2020,
  {DOI}: 10.1109/IROS45743.2020.9341396.

\bibitem{lee22ifac}
M.-H. Lee and J.-S. Liu,
  ``\href{https://doi.org/10.1016/j.ifacol.2022.07.622}{Single Swept Volume
  Reconstruction by Signed Distance Function Learning: A feasibility study
  based on implicit geometric regularization},'' \emph{IFAC-PapersOnLine},
  vol.~55, no.~15, pp. 142--147, 2022, {C}onf. on Intelligent Control and
  Automation Sciences {(ICONS)}.

\bibitem{gropp20icml}
A.~Gropp, L.~Yariv, N.~Haim, M.~Atzmon, and Y.~Lipman,
  ``\href{https://dl.acm.org/doi/pdf/10.5555/3524938.3525293}{Implicit
  Geometric Regularization for Learning Shapes},'' in \emph{Proc.~of the
  Int.~Conf.~on Machine Learning {(ICML)}}, 2020, {DOI}:
  10.5555/3524938.3525293.

\bibitem{michaux23rss}
J.~B. Michaux, Y.~Kwon, Q.~Chen, and R.~Vasudevan,
  ``\href{https://doi.org/10.15607/RSS.2023.XIX.062}{Reachability-based
  Trajectory Design with Neural Implicit Safety Constraints},'' in
  \emph{Proc.~of Robotics: Science and Systems ({RSS})}, 2023, {DOI}:
  10.15607/RSS.2023.XIX.062.

\bibitem{liu22iros}
P.~Liu, K.~Zhang, D.~Tateo, S.~Jauhri, J.~Peters, and G.~Chalvatzaki,
  ``\href{https://doi.org/10.1109/IROS47612.2022.9981456}{Regularized Deep
  Signed Distance Fields for Reactive Motion Generation},'' in \emph{Proc.~of
  the {IEEE}/{RSJ} Int.~Conf.~on Intelligent Robots and Systems ({IROS})},
  2022, {DOI}: 10.1109/IROS47612.2022.9981456.

\bibitem{koptev23ral}
M.~Koptev, N.~Figueroa, and A.~Billard,
  ``\href{https://doi.org/10.1109/LRA.2022.3227860}{Neural Joint Space Implicit
  Signed Distance Functions for Reactive Robot Manipulator Control},''
  \emph{{IEEE} Robotics and Automation Letters}, vol.~8, no.~2, pp. 480--487,
  2023.

\bibitem{li23signedDist}
Y.~Li, Y.~Zhang, A.~Razmjoo, and S.~Calinon,
  ``\href{https://doi.org/10.48550/arXiv.2307.00533}{Learning Robot Geometry as
  Distance Fields: Applications to Whole-body Manipulation},'' 2023, arXiv
  preprint, {DOI}: 10.48550/arXiv.2307.00533.

\bibitem{resnet}
K.~He, X.~Zhang, S.~Ren, and J.~Sun,
  ``\href{https://doi.org/10.1109/CVPR.2016.90}{Deep Residual Learning for
  Image Recognition},'' in \emph{Proc.~of the {IEEE} Conf.~on Computer Vision
  and Pattern Recognition {(CVPR)}}, 2016, pp. 770--778, {DOI}:
  10.1109/CVPR.2016.90.

\bibitem{park19cvpr}
J.~J. Park, P.~Florence, J.~Straub, R.~Newcombe, and S.~Lovegrove,
  ``\href{https://doi.org/10.1109/CVPR.2019.00025}{{DeepSDF}: Learning
  Continuous Signed Distance Functions for Shape Representation},'' in
  \emph{Proc.~of the {IEEE} Conf.~on Computer Vision and Pattern Recognition
  {(CVPR)}}, 2019, {DOI}: 10.1109/CVPR.2019.00025.

\bibitem{gpytoolbox}
S.~Sell\'{a}n, O.~Stein, \emph{et~al.}, ``{G}pytoolbox: A python geometry
  processing toolbox,'' 2023,
  \href{https://gpytoolbox.org}{https://gpytoolbox.org}.

\bibitem{libigl}
A.~Jacobson, D.~Panozzo, \emph{et~al.}, ``{libigl}: A simple {C++} geometry
  processing library,'' 2023,
  \href{https://libigl.github.io}{https://libigl.github.io}.

\bibitem{fcl}
J.~Pan, S.~Chitta, and D.~Manocha,
  ``\href{https://doi.org/10.1109/ICRA.2012.6225337}{{FCL}: A General Purpose
  Library for Collision and Proximity Queries},'' in \emph{Proc.~of the {IEEE}
  Int.~Conf.~on Robotics and Automation ({ICRA})}, 2012, pp. 3859--3866, {DOI}:
  10.1109/ICRA.2012.6225337.

\bibitem{icra24sweptVolDataset}
D.~Joho, J.~Schwinn, and K.~Safronov,
  ``\href{https://doi.org/10.5281/zenodo.10638607}{Data, Models, and Code for:
  Neural Implicit Swept Volume Models for Fast Collision Detection},'' Dataset
  on zenodo.org, 2024, {DOI}: 10.5281/zenodo.10638607.

\end{thebibliography}

\end{document}